\documentclass[11pt]{article}
\PassOptionsToPackage{table}{xcolor}
\usepackage{booktabs}
\usepackage{multirow}
\usepackage{array}
\usepackage{graphicx}
\usepackage[table]{xcolor}
\usepackage[]{acl}
\usepackage{times}
\usepackage{latexsym}
\usepackage[T1]{fontenc}
\usepackage[utf8]{inputenc}
\usepackage{microtype}
\usepackage{inconsolata}
\usepackage{graphicx}
\usepackage{amsmath}
\usepackage{subfigure}
\usepackage{amssymb}
\usepackage{dsfont}
\usepackage{float}

\title{TokenPilot: Cache-Efficient Context Management for LLM Agents}
\author{
  Buqiang Xu\textsuperscript{1}\thanks{ \ \ Equal contribution.},
  Zirui Xue\textsuperscript{1}\footnotemark[1],
  Dianmou Chen\textsuperscript{2}\footnotemark[1],
  Chenyang Fu\textsuperscript{3}\footnotemark[1],
  Chiyu Wu\textsuperscript{4}\footnotemark[1], \\
  \textbf{Caiying Huang\textsuperscript{3}\footnotemark[1], 
  Chen Jiang\textsuperscript{1}, 
  Jizhan Fang\textsuperscript{1},
  Xinle Deng\textsuperscript{1},
  Yijun Chen\textsuperscript{1}}, \\
  \textbf{Yunzhi Yao\textsuperscript{1}, 
  Xuehai Wang\textsuperscript{1},
  Jin Shang\textsuperscript{4},
  Gong Yu\textsuperscript{4}, 
  Ningyu Zhang\textsuperscript{1}}\thanks{ \ \ Corresponding author.}
  \\
  \\
  \textsuperscript{1}Zhejiang University
  \quad
  \textsuperscript{2}University of Electronic Science and Technology of China\\
  \textsuperscript{3}Xi'an University of Electronic Science and Technology \quad
  \textsuperscript{4}HomologyAI \\
}

\begin{document}
\maketitle
\begin{abstract}
As LLM agents are deployed in long-horizon sessions, context accumulation drives up inference costs. Existing approaches utilize text pruning or dynamic memory eviction to minimize token footprints; however, their unconstrained sequence mutations alter layouts, introducing prefix mismatches and cache invalidation. This reveals a critical trade-off between text sparsity and prompt cache continuity. To address this, we present \textbf{TokenPilot}, a dual-granularity context management framework. Globally, \textit{Ingestion-Aware Compaction} acts as a framework harness to stabilize prompt prefixes and eliminate open-world environmental noise at the ingestion gate. Locally, \textit{Lifecycle-Aware Eviction} monitors the ongoing residual utility of context segments, enforcing a conservative batch-turn schedule to offload content segments only when task relevance expires. Experiments on \texttt{PinchBench} and \texttt{Claw-Eval} under both isolated and continuous modes demonstrate that TokenPilot reduces costs by \textbf{61\%} and \textbf{56\%} in isolated mode, and \textbf{61\%} and \textbf{87\%} in continuous mode, while maintaining competitive performance compared to prior systems\footnote{TokenPilot has been integrated into LightRSI at \url{https://github.com/zjunlp/LightRSI}.}.  
\end{abstract}

\section{Introduction} \label{Introduction}
The paradigm of large language models has shifted from conversational assistants~\cite{llm_as_conversational_assistant} to stateful execution controllers~\cite{claudecode,codex,openclaw} orchestrating complex tools~\cite{toolathlon,terminalbench}, file systems~\cite{swebench}, and cross-application workflows~\cite{webarena}. 
Consequently, the core challenge of agent design has transitioned to real-world operational reliability~\cite{Claw-Eval,pinchbench}. 
However, continuous multi-turn interactions inevitably accumulate verbose execution traces, rapidly inflating sequence lengths and escalating per-turn inference costs. 
Managing this context growth is thus an essential prerequisite for sustainable real-world deployment~\cite{memorysurvey, ContextEngineeringSurvey}.
 
\begin{figure}[!t]
    \centering
    \includegraphics[width=0.9\linewidth]{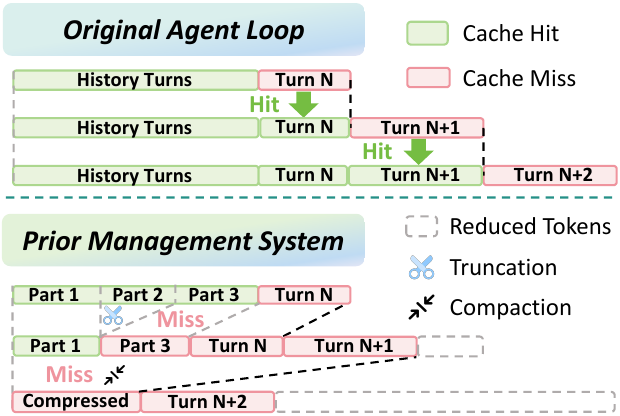}
    \caption{Comparison of cache alignment behaviors. While the \textit{Original Agent Loop} maintains continuous layouts to achieve cumulative \textcolor{green!60!black}{\textbf{cache hits}}, previous management systems execute text truncation or compaction that mutates input boundaries, inadvertently triggering severe backend KV \textcolor{red!70!black}{\textbf{cache misses}}.}
    \vspace{-5mm}
    \label{fig:intro}
\end{figure}

The research community has primarily approached this challenge from a content-reduction perspective. 
Initial efforts focus on static text compression, pruning low-utility tokens~\cite{llmlingua, llmlingua2} or sentences~\cite{SelectiveContext} before prompt transmission. 
For dynamic execution traces, existing architectures implement context folding~\cite{agentfold, agentswing}, or demand paging~\cite{Pichay} to condense intermediate reasoning backbones~\cite{memobrain} and offload continuous trajectories to external storage~\cite{MemOS,evermemos}. 
 
Despite their success in textual compaction, these methods introduce a fundamental trade-off between prompt reduction and hardware cache efficiency~\cite{vllm, sglang}. 
As shown in Figure~\ref{fig:intro}, while aggressively truncating or shifting context pages minimizes per-turn token counts, this constant layout mutation shatters prompt prefix continuity. 
The resulting hardware pre-fill penalties and cache invalidations ultimately override any financial savings from text reduction. 
\textbf{We argue that an effective framework must fundamentally reconcile text-level sparsity with hardware cache alignment.} 
To achieve this design synergy, the deployment system must simultaneously safeguard physical prefix continuity during observation ingestion and defer structural memory eviction until a trajectory's residual utility thoroughly expires.

Building on this insight, we present \textbf{TokenPilot}, a dual-granularity context management framework that reconciles sequence reduction with prompt cache alignment. 
At the global level, \textit{Ingestion-Aware Compaction} acts as a deterministic harness to optimize the layout at the initial warm-up phase rather than retroactively compressing an existing cache. 
Specifically, it neutralizes volatile runtime variables via stable placeholders and shifts tool definitions downstream to secure a byte-identical prompt prefix from the first turn, while concurrently stripping structural noise from incoming tool responses before ingestion.
At the local level, \textit{Lifecycle-Aware Eviction} monitors active execution trajectories online by evaluating their dynamic residual utility. 
Rather than executing frequent, disruptive memory paging, the eviction pass remains strictly conservative, deferring structural purge until the segment's residual value thoroughly expires to safeguard context continuity. 

Evaluated on \texttt{PinchBench} and \texttt{Claw-Eval} under commercial pricing structures, TokenPilot dramatically reduces total inference monetary expenditures by \textbf{61\%} and \textbf{56\%} in isolated mode, and \textbf{61\%} and \textbf{87\%} in continuous mode, while successfully maintaining competitive task performance. 
\section{Background} \label{Background}
\paragraph{Task Settings.}
We consider an agent processing a sequence of tasks $\mathcal{S} = \{t_1, t_2, \ldots, t_n\}$. Each task generates a trajectory of instructions, reasoning traces, tool calls, and responses, which accumulate into the session context $\mathcal{C}$. We evaluate under two modes: \textit{isolated mode}, where the context $\mathcal{C}$ is reset at each task boundary, and \textit{continuous mode}, where histories persist across the entire sequence.

\paragraph{Optimization Objective.}
A context management framework $\mathcal{M}$ transforms the raw history $\mathcal{C}$ into an optimized runtime context $\mathcal{C}' = \mathcal{M}(\mathcal{C})$. The objective is to maximize the ratio of context utility to maintenance cost:
\begin{equation}
\max_{\mathcal{M}} \frac{\sum_{m \in \mathcal{C}'} \hat{U}(m \mid \mathcal{C}')}{\mathcal{K}(\mathcal{C}')}
\end{equation}
Here, context utility quantifies the necessity of context tokens for guiding downstream reasoning and tool execution, where $\hat{U}(m \mid \mathcal{C}')$ estimates the marginal contribution of message $m$ to subsequent agent actions. The serving cost $\mathcal{K}(\mathcal{C}')$ is governed by the backend KV prompt-caching mechanism:
\begin{equation}
\mathcal{K}(\mathcal{C}') = \alpha \cdot |\mathcal{C}'_{\text{hit}}| + |\mathcal{C}'_{\text{miss}}|
\end{equation}
where $\mathcal{C}'_{\text{hit}}$ and $\mathcal{C}'_{\text{miss}}$ denote tokens served from the cache at a discounted cost rate $\alpha \ll 1$ and those incurring full pre-fill cost, respectively, subject to the length alignment constraint $|\mathcal{C}'| = |\mathcal{C}'_{\text{hit}}| + |\mathcal{C}'_{\text{miss}}|$.
\section{TokenPilot} \label{Method}

\begin{figure*}[!t]
    \centering
    \includegraphics[width=1.0\linewidth]{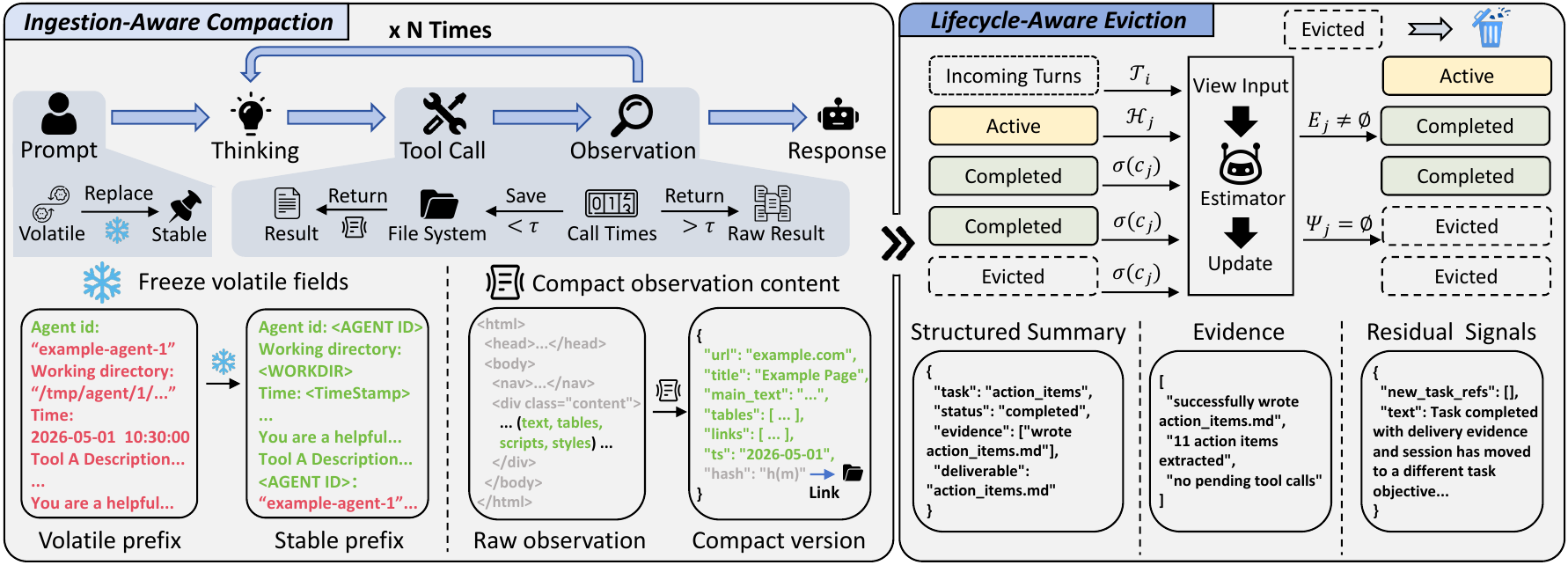}
    \caption{The system architecture of TokenPilot, featuring Ingestion-Aware Compaction at the global framework harness level and Lifecycle-Aware Eviction at the local context sequence level.}
    \label{fig:overview}
\end{figure*}

\subsection{Overall}
We propose \textbf{TokenPilot}, a dual-granularity framework that addresses the context optimization objective across two complementary operational levels.
At the \textit{global} framework level, \textit{Ingestion-Aware Compaction} (\S\ref{sec:conditioning}) acts as a deterministic harness at the ingestion boundary, standardizing layouts and purifying incoming messages to optimize the sequence during the initial cache warm-up phase.
At the \textit{local} sequence level, \textit{Lifecycle-Aware Eviction} (\S\ref{sec:eviction}) dynamically monitors the residual utility of active trajectories, enforcing a conservative batch-turn schedule to purge segments only when their task-level utility has thoroughly expired.
 
\subsection{Global Ingestion-Aware Compaction}
\label{sec:conditioning}
Ingestion-Aware Compaction acts as a framework harness to optimize sequence layout at the ingestion boundary. 
Based on the interaction loop, we partition the message space into two functional categories. 
Let $\Omega_{\text{int}}$ denote \textit{internal intentional messages} generated natively by the system or model, encompassing task prompts, thinking traces, tool calls, and final responses, which naturally possess high intrinsic utility density. 
Conversely, let $\Omega_{\text{env}}$ denote \textit{open-world environmental feedback} from unmanaged tools, which inherently suffer from extensive structural clutter. 
The marginal utility density for an incoming message $m$ is formalized as:
\begin{equation}
\hat{U}(m) = \begin{cases} 
1 & m \in \Omega_{\text{int}} \\ 
\max(\gamma(m), \mathbb{G}(m)) & m \in \Omega_{\text{env}}
\end{cases} 
\end{equation}
where $\gamma(m) \le 1$ is the base environmental feedback utility density, and $\mathbb{G}(m) = \mathds{1}[f(h(m)) > \tau]$ is an ingestion gate. When the access frequency $f(h(m))$ of a content hash exceeds threshold $\tau$, $\mathbb{G}(m)$ switches to $1$, fully upgrading the density to restore comprehensive content delivery.
 
\paragraph{Prefix Stabilization.} 
Cross-task KV cache reuse is frequently disrupted by runtime-volatile fields within $\Omega_{\text{int}}$ that introduce position jitter and prefix mismatches. 
TokenPilot implements a canonicalization operator to secure a byte-identical prompt prefix from the first turn:
\begin{equation}
\phi(m^{(t)}) = \phi(m^{(t+1)}) \implies \mathcal{C}'_{\text{prefix}} \subseteq \mathcal{C}'_{\text{hit}}
\end{equation}
where $\phi$ intercepts internal messages at the harness level and substitutes volatile runtime markers with static placeholders. 
By preserving physical continuity across tasks, this operator directly eliminates full-cost pre-fill penalties.

\paragraph{Observation Reduction.}
For environmental messages $m \in \Omega_{\text{env}}$ where $\mathbb{G}(m) = 0$, TokenPilot applies deterministic reduction passes at the ingestion gate to lift token utility. 
The transformation and its reliable fallback loop are formalized as:
\begin{equation}
m_{\text{ingest}} = \kappa(m), \quad \mathcal{A}[h(m)] \leftarrow m
\end{equation}
where $\kappa(m)$ denotes the compacted structural preview stored in working memory, and $\mathcal{A}$ represents an external artifact registry indexed by content hash $h(m)$ to ensure total operational safety. 
If the compressed message lacks critical signals during execution, the agent harness invokes a lightweight recovery tool to dynamically recall the full payload $\mathcal{A}[h(m)]$, automatically upgrading the status to disable subsequent truncation for that path.

\subsection{Local Lifecycle-Aware Eviction}
\label{sec:eviction}
At the local sequence level, Lifecycle-Aware Eviction dynamically regulates the historical retention window based on a segment's ongoing task utility. 
To safeguard physical cache continuity and prevent disruptive turn-by-turn memory paging, TokenPilot tracks each context segment $c_j$ through three progressive states $s_j \in \{\textit{active}, \textit{completed}, \textit{evictable}\}$ maintained in a framework registry $\mathcal{R}$. 
The marginal utility of a segment is formalized as: 
\begin{equation}
\hat{U}(c_j) = \begin{cases} 
1 & s_j = \textit{active} \\
\mathds{1}[\Psi_j \neq \emptyset] & s_j = \textit{completed} \\
0 & s_j = \textit{evictable}
\end{cases}
\end{equation}
where $\Psi_j$ represents the quantified \textit{residual utility} of the segment. Under this formulation, a segment that has concluded its execution path is not immediately truncated; it transitions to a conservative \textit{completed} state, retaining its physical cache slots as long as its residual relevance to ongoing interactions remains non-zero ($\Psi_j \neq \emptyset$).
 
\definecolor{VanillaColor}{RGB}{235,235,235}
\definecolor{StaticCompressionColor}{RGB}{255,228,225}
\definecolor{DynamicMgmtColor}{RGB}{220,237,255}
\definecolor{ProceduralMemoryRetrievalColor}{RGB}{230,245,225}
\definecolor{EpisodicMemoryRetrievalColor}{RGB}{235,225,245}
\definecolor{SemanticMemoryRetrievalColor}{RGB}{246,226,193}

\begin{table*}[!htbp]
\centering
\resizebox{\linewidth}{!}{%
\begin{tabular}{l|c|ccccccccccc|cc|c|c}
\toprule[1pt]
\multirow{2}{*}{\textbf{Method}} & \multirow{2}{*}{\textbf{Overall $\uparrow$}} 
& \multicolumn{11}{c|}{\textbf{Performance by Category $\uparrow$}} 
& \multicolumn{2}{c|}{\textbf{Input Tokens (M)}} 
& \multirow{2}{*}{\textbf{Output (M)}}
& \multirow{2}{*}{\textbf{Cost (\$) $\downarrow$}} \\
\cmidrule(lr){3-13} \cmidrule(lr){14-15}
& & \textbf{Prod} & \textbf{Res} & \textbf{Write} & \textbf{Code} & \textbf{Anal} 
& \textbf{CSV} & \textbf{Log} & \textbf{Meet} & \textbf{Mem} & \textbf{Skill} & \textbf{Integ} 
& \textbf{Cache Hit} & \textbf{Cache Miss} & & \\
\midrule
\multicolumn{17}{l}{\textbf{\textit{Isolated Mode}}} \\
\midrule
 Vanilla & \underline{80.5} & 87.2 & 68.7 & 84.1 & 86.0 & 75.1 & 83.0 & 94.7 & 81.4 & 86.5 & 70.3 & 55.3 & 6.184 & 8.753 & 0.285 & 8.31 \\
 LLMLingua-2      & 76.9 & 89.3 & 64.0 & 82.1 & 86.9 & 80.8 & 79.6 & 84.4 & 66.3 & 85.0 & 79.6 & 72.1 & 14.241 & 3.975 & 0.384 & 5.78 \\
 SelectiveContext & 76.5 & 88.5 & 64.5 & 73.0 & 83.7 & 82.6 & 81.1 & 92.8 & 63.3 & 86.9 & 82.8 & 77.2 & 11.273 & 4.642 & 0.324 & 5.79 \\
 LCM         & 77.8 & 90.1 & 64.9 & 79.6 & 85.4 & 81.3 & 81.0 & 87.1 & 67.5 & 85.0 & 81.7 & 80.6 & 16.018 & 3.064 & 0.356 & 5.10 \\
 Pichay      & 78.9 & 85.4 & 58.9 & 71.8 & 79.0 & 88.3 & 79.8 & 83.6 & 84.0 & 91.3 & 69.8 & 63.3 & 6.717 & 3.333 & 0.238 & 4.07 \\
 Summary     & 79.5 & 80.7 & 66.3 & 83.5 & 77.9 & 82.1 & 87.5 & 77.2 & 81.3 & 92.5 & 67.2 & 54.4 & 12.303 & 3.009 & 0.296 & 4.51\\
 MemoBrain   & 78.1 & 86.8 & 62.1 & 88.9 & 85.7 & 82.6 & 88.3 & 85.4 & 63.6 & 92.5 & 76.1 & 69.7 & 10.200 & 2.107 & 0.233 & \underline{3.36} \\
 AgentSwing  & 78.4 & 89.8 & 71.9 & 80.2 & 79.5 & 83.5 & 80.8  & 83.7 & 77.9 & 92.5 & 65.7 & 35.0 & 4.534 & 7.129 & 0.241 & 6.77 \\
 Keep-Last-N & 80.4 & 86.0 & 70.0 & 82.4 & 80.1 & 77.6 & 78.3 & 91.5 & 84.3 & 92.5 & 70.1 & 87.8 & 12.813 & 2.657 & 0.291 & 4.26 \\
 MemOS  & 79.4 & 84.2 & 54.4 & 83.1 & 82.3 & 78.2 & 81.1 & 97.2 & 77.6 & 92.5 & 85.9 & 80.2 & 29.018 & 4.573 & 0.492 & 7.81 \\
 \midrule
 \textbf{TokenPilot} & \textbf{81.0} & 89.0 & 71.2 & 80.0 & 72.6 & 88.9 & 85.3 & 95.2 & 79.4 & 95.0 & 95.2 & 58.0 & 8.893 & 1.933 & 0.244 & \textbf{3.22} \\
\midrule
\multicolumn{17}{l}{\textbf{\textit{Continuous Mode}}} \\
\midrule
 Vanilla & 79.2 & 83.5 & 58.4 & 86.8 & 80.0 & 78.5 & 87.8 & 94.6 & 77.6 & 95.0 & 55.8 & 83.6 & 25.015 & 5.943 & 0.202 & 7.24 \\
 LLMLingua-2      & 73.8 & 85.8 & 58.4 & 80.3 & 74.3 & 79.6 & 82.8 & 84.2 & 63.4 & 90.0 & 79.1 & 83.6 & 20.574 & 2.183 & 0.194 & 4.06 \\
 SelectiveContext & 74.0 & 85.4 & 64.2 & 83.1 & 75.4 & 78.8 & 77.3 & 91.2 & 62.2 & 89.5 & 71.0 & 80.3 & 25.475 & 2.608 & 0.196 & 4.75 \\
 LCM         & 77.0 & 88.1 & 63.2 & 90.1 & 75.7 & 78.5 & 85.4 & 88.9 & 65.1 & 82.8 & 80.8 & 78.2 & 18.708 & 2.417 & 0.222 & 4.21 \\
 Pichay      & 76.5 & 88.0 & 66.7 & 76.2 & 81.0 & 77.6 & 83.5 & 84.2 & 67.6 & 100.0 & 63.8 & 75.3 & 11.698 & 6.874 & 0.260 & 7.20 \\
 Summary     & 78.4 & 89.1 & 64.4 & 73.8 & 82.9 & 69.6 & 81.6 & 93.6 & 80.3 & 95.0 & 61.7 & 75.3 & 20.687 & 6.249 & 0.196 & 7.12 \\
 MemoBrain   & 78.0 & 87.7 & 65.0 & 85.5 & 84.9 & 75.9 & 81.0 & 89.0 & 72.3 & 90.3 & 86.6 & 84.7 & 12.917 & 2.283 & 0.232 & \underline{3.73} \\
 AgentSwing  & 78.5 & 86.3 & 67.3 & 89.0 & 79.1 & 82.4 & 87.4 & 68.1 & 72.4 & 93.8 & 61.7 & 83.8 & 12.680 & 5.476 & 0.314 & 6.47 \\
 Keep-Last-N & 79.1 & 86.3 & 67.0 & 87.8 & 87.0 & 77.0 & 85.4 & 77.3 & 75.9 & 95.0 & 56.8 & 75.1 & 18.117 & 4.481 & 0.209 & 5.66 \\
 MemOS & \underline{80.9} & 87.5 & 59.0 & 85.4 & 87.1 & 82.0 & 81.0 & 95.0 & 78.1 & 92.5 & 87.4 & 84.1 & 30.859 & 8.939 & 0.308 & 10.41 \\
 \midrule
 \textbf{TokenPilot} & \textbf{81.3} & 76.7 & {76.9} & {90.6} & {84.1} & {86.0} & {85.6} & {89.1} & 73.6 & {95.0} & {77.2} & 80.1 & 8.551 & 1.549 & 0.219 & \textbf{2.79} \\
\bottomrule[1pt]
\end{tabular}
} 
\caption{Performance and resource consumption comparison on \texttt{PinchBench} under isolated and continuous modes. $\uparrow$: larger is better; $\downarrow$: smaller is better. \textbf{Best results} in bold, \underline{second-best} underlined. \textbf{Input Tokens} are decomposed into \textbf{Cache Hit} and \textbf{Cache Miss} tokens, reflecting prefix stability and reuse efficiency.  Category abbreviations: Prod=Productivity, Res=Research, Write=Writing, Code=Coding, Anal=Analysis, CSV=CSV Analysis, Log=Log Analysis, Meet=Meeting Analysis, Mem=Memory, Skill=Skills, Integ=Integrations.}
\label{tab:pinchbench}
\end{table*} 
\paragraph{Context State Estimation and Execution.}
To suppress spurious state transitions, an online model-based estimator $\mathcal{E}$ is triggered conservatively in stable batches of $B$ turns rather than at every execution step. 
For the $i$-th batch, the estimator ingests a compressed historical view $\mathcal{V}_i$ to compute state updates over each segment:
\begin{equation}
\Delta\mathcal{R}_i^{(j)} = \langle E_j,\, \Psi_j \rangle = \mathcal{E}(\mathcal{V}_i, \mathcal{R}_{i-1})
\end{equation}
where $E_j$ denotes explicit resolution evidence showing the sub-task has achieved its objective, and $\Psi_j$ represents residual utility signals extracted from dependency patterns. 
TokenPilot enforces a gated pipeline to transition these lifecycle states:
\begin{equation}
\textit{active} \xrightarrow{E_j \neq \emptyset} \textit{completed} \xrightarrow{\Psi_j = \emptyset} \textit{evictable}
\end{equation}
The registry executes strict system validation, updating via $\mathcal{R}_i \leftarrow \mathcal{R}_{i-1} \oplus \Delta\mathcal{R}_i$ only for valid transitions. 
Once a segment drops to $s_j = \textit{evictable}$, its utility decays to zero, and the framework executes a single-pass structural purge to construct the optimized context window $\mathcal{C}'$:
\begin{equation}
\mathcal{C}' = \{ m \in \mathcal{C} \mid s_{j(m)} \neq \textit{evictable} \}
\end{equation}
where $j(m)$ maps message $m$ to its segment index. 
This batch-gated execution guarantees that eviction remains highly restrained, maximizing cache continuity by eliminating volatile text mutation.

To operationalize this, the estimator $\mathcal{E}$ is instantiated via \texttt{Qwen3.5-35B-A3B} as a lightweight, zero-shot validator, incurring negligible overhead; for instance, its total operational cost across the continuous \texttt{PinchBench} stream is less than \textbf{\$0.03}.
\section{Experiments} \label{Experiments}
\subsection{Experimental Setup}
\label{sec:setup}

\paragraph{Benchmarks and Metrics.}
We evaluate TokenPilot on \texttt{PinchBench} and \texttt{Claw-Eval} across both \textit{isolated} and \textit{continuous} modes (see Appendix~\ref{Appendix:Dataset} for dataset statistics). We track task accuracy alongside actual monetary expenditures. To ensure empirical fidelity, all cache hit and miss token counts are gathered directly from the explicit metadata fields returned by the provider APIs, eliminating client-side estimation errors. Joint scoring formulas and pricing tiers are detailed in Appendix~\ref{Appendix:Metrics}.
 
\paragraph{Implementation Details.} We compare TokenPilot against compression methods (LLMLingua-2, SelectiveContext, Keep-Last-N) and dynamic paging or summarization approaches (Summary, LCM, Pichay, MemoBrain, AgentSwing, MemOS). 
All evaluated methods utilize \texttt{GPT-5.4-mini} as the agent backbone. 
Detailed hyperparameter configurations for all baselines are documented in Appendix~\ref{appendix:baseline}. 
The exact execution thresholds, model assignments, and system prompts for \texttt{TokenPilot} are detailed in Appendix~\ref{appendix:Implementation}.

\definecolor{VanillaColor}{RGB}{235,235,235}
\definecolor{StaticCompressionColor}{RGB}{255,228,225}
\definecolor{DynamicMgmtColor}{RGB}{220,237,255}
\definecolor{MemoryRetrievalColor}{RGB}{230,245,225}
\definecolor{OursColor}{RGB}{235,225,245}

\begin{table*}[!htbp]
\centering
\resizebox{\linewidth}{!}{%
\begin{tabular}{l|c|ccccccccccc|cc|c|c}
\toprule[1pt]
\multirow{2}{*}{\textbf{Method}} & \multirow{2}{*}{\textbf{Overall $\uparrow$}}
& \multicolumn{11}{c|}{\textbf{Performance by Category $\uparrow$}}
& \multicolumn{2}{c|}{\textbf{Input Tokens (M)}}
& \multirow{2}{*}{\textbf{Output (M)}}
& \multirow{2}{*}{\textbf{Cost (\$) $\downarrow$}} \\
\cmidrule(lr){3-13} \cmidrule(lr){14-15}
& & \textbf{Wkfl} & \textbf{Ops} & \textbf{Fin} & \textbf{Off} & \textbf{Comm}
& \textbf{Prod} & \textbf{Oprn} & \textbf{Safe} & \textbf{Term} & \textbf{MM} & \textbf{Oth}
& \textbf{Cache Hit} & \textbf{Cache Miss} & & \\
\midrule
\multicolumn{17}{l}{\textbf{\textit{Isolated Mode}}} \\
\midrule
 Vanilla          & \textbf{64.5} & 65.4 & 70.8 & 45.7 & 44.4 & 73.2 & 70.9 & 77.7 & 74.0 & 56.8 & 41.0 & 69.2  & 9.429 & 4.637 & 0.216 & 5.16  \\
 LLMLingua-2 & 61.9 & 58.7 & 67.5 & 57.6 & 43.3 & 62.9 & 70.1 & 62.4 & 61.0 & 49.6 & 44.0 & 75.2 & 8.169 & 4.043 & 0.182 & 4.44 \\
 SelectiveContext  & 60.7 & 59.1 & 68.2 & 46.3 & 36.9 & 61.5 & 75.5 & 59.2 & 67.2 & 53.1 & 44.0 & 74.7 & 8.271 & 3.862 & 0.181 & 4.31 \\
 LCM & 61.2 & 59.0 & 67.3 & 51.1 & 47.7 & 65.9 & 76.6 & 58.4 & 58.6 & 51.4 & 41.5 & 72.2 & 9.776 & 3.543 & 0.172 & 4.17 \\
 Pichay       & 59.3 & 57.3 & 62.1 & 38.2 & 39.4 & 68.5 & 65.0 & 91.6 & 64.1 & 25.6 & 55.0 & 76.5 & 4.648 & 3.944  & 0.186 & 4.14 \\
 Summary      & 62.0 & 70.0 & 71.0 & 32.2 & 20.6 & 80.0 & 68.5 & 82.8 & 49.2 & 20.0 & 41.0 & 71.4 & 2.935 & 2.871 & 0.174 & 3.16 \\
 MemoBrain  & 58.0 & 64.5 & 60.5 & 26.1 & 37.6 &56.1 & 59.9& 71.0 & 63.4& 20.0 & 41.0& 75.3& 18.182 & 5.118 & 0.332 & 6.69 \\
 AgentSwing   & 60.9 & 64.2 & 66.5 & 44.1 & 45.7 & 67.8 & 52.8 & 85.8 & 57.2 & 25.6 & 53.6 & 68.8 & 4.580 & 3.585 & 0.194 & 3.91 \\
 Keep-Last-N  & 61.8 & 67.1 & 73.8 & 44.7 & 21.6 & 54.5 & 63.6 & 86.2 & 38.4 & 39.4 & 55.0 & 69.1 & 4.229 & 1.845 & 0.186 & \underline{2.54} \\
 MemOS & 61.6 & 64.7 & 74.2 & 40.9 & 25.2 & 71.2 & 32.0 & 73.6 & 80.2 & 20.0 & 56.2 & 74.6 & 12.582 & 2.709 & 0.363 & 4.61 \\
 \midrule
 \textbf{TokenPilot} & \underline{63.1} & 68.1 & 75.4 & 47.0 & 22.3 & 71.8 & 65.0 & 72.0 & 47.8 & 37.0 & 45.6 & 69.9 & 4.436 & 1.154 & 0.239 & \textbf{2.27} \\
\midrule
\multicolumn{17}{l}{\textbf{\textit{Continuous Mode}}} \\
\midrule
 Vanilla          & \textbf{63.4} & 70.8 & 80.3 & 26.7 & 27.8 & 62.2 & 73.4 & 78.4 & 63.6 & 20.0 & 41.0 & 69.6 & 709.845 & 21.981 & 2.622 & 81.52  \\
 LLMLingua-2      & 59.0 & 58.7 & 71.3 & 34.8 & 30.6 & 61.9 & 65.3 & 77.6 & 64.6 & 20.0 & 41.0 & 72.4 & 575.654 & 37.197 & 2.630 & 82.91 \\
 SelectiveContext  & 56.5 & 58.1 & 71.6 & 21.8 & 21.2 & 54.7 & 74.0 & 57.7 & 66.4 & 20.0 & 41.0 & 72.3 & 437.114 & 48.678 & 2.754 & 81.69 \\
 LCM          & 61.4 & 66.8 & 69.0 & 38.3 & 29.5 & 63.3 & 74.9 & 66.6 & 67.3 & 20.0 & 41.0 & 72.7 & 383.007 & 28.714 & 2.691 & 62.37 \\
 Pichay       & 61.0 & 69.5 & 63.8 & 40.3 & 24.0 & 63.1 & 67.0 & 94.1 & 52.5 & 21.6 & 41.0 & 71.0 & 97.431 & 63.510 & 1.046 & 59.65 \\
 Summary      & 61.6 & 63.6 & 74.5 & 35.3 & 20.6 & 55.5 & 70.1 & 87.1 & 66.1 & 69.0 & 42.6 & 66.9 & 59.772 & 10.143 & 1.001 & 16.59 \\
 MemoBrain   & 57.9 & 65.9 & 55.0 & 24.9 & 36.7 & 47.8& 73.5 & 64.2 & 60.6 & 20.0 & 38.4 & 81.6 & 47.497 & 13.990 & 1.134 & 19.16 \\
 AgentSwing   & \underline{62.2} & 67.6 & 66.5 & 48.6 & 36.8 & 70.0 & 63.8 & 90.7 & 31.7 & 22.4 & 41.0 & 72.8 & 53.776 & 10.027 & 0.907 & 15.63 \\
 Keep-Last-N  & 60.7 & 65.3 & 74.0 & 35.5 & 20.8 & 54.1 & 73.6 & 91.9 & 35.7 & 59.5 & 42.4 & 64.7 & 44.812 & 9.106 & 0.780 & \underline{13.70} \\
 MemOS & 57.7 & 55.9 & 65.0 & 56.3 & 22.2 & 44.8 & 64.6 & 68.8 & 89.0 & 20.0 & 39.6 & 71.5 & 49.742 & 25.432 & 0.293 & 24.12 \\
 \midrule
 \textbf{TokenPilot} & 60.8 & 58.8 & 61.8 & 52.5 & 32.1 & 64.2 & 57.3 & 89.2 & 65.8 & 76.8 & 45.2 & 70.9 & 21.430 & 9.928 & 0.338 & \textbf{10.58} \\ 
\bottomrule[1pt]
\end{tabular}
}
\caption{Performance and resource consumption comparison on \texttt{Claw-Eval} under isolated and continuous modes. $\uparrow$: larger is better; $\downarrow$: smaller is better. \textbf{Best results} in bold, \underline{second-best} underlined. \textbf{Input Tokens} are decomposed into \textbf{Cache Hit} and \textbf{Cache Miss} tokens. Category abbreviations: Wkfl=Workflow, Ops=Ops, Fin=Finance, Off=Office QA, Comm=Communication, Prod=Productivity, Oprn=Operations, Safe=Safety, Term=Terminal, MM=Multimodal, Oth=Others.}
\label{tab:claw-eval}
\end{table*}
\begin{table}[t]
\centering
\setlength{\tabcolsep}{8pt}
\resizebox{\linewidth}{!}{%
\begin{tabular}{l|cc|ccc}
\toprule[1pt]
\textbf{Method} & \textbf{Overall} & \textbf{Cost (\$)} & \textbf{Hit (M)} & \textbf{Miss (M)} & \textbf{Output (M)} \\
\midrule
Vanilla                  & 79.2 & 7.24 & 25.015 & 5.943 & 0.202 \\
+ Global Level   & 79.9 & 4.22 & 26.716 & 1.589 & 0.227 \\
+ Local Level     & 81.3 & 2.79 &  8.551 & 1.549 & 0.219 \\
\bottomrule[1.2pt]
\end{tabular}
}
\caption{Progressive ablation of TokenPilot components on \texttt{PinchBench} in continuous mode. ``Hit'' and ``Miss'' denote the token counts for cache hits and cache misses.}
\label{tab:ablation_progressive}
\end{table}
\begin{figure}[t]
    \centering
    \includegraphics[width=\linewidth]{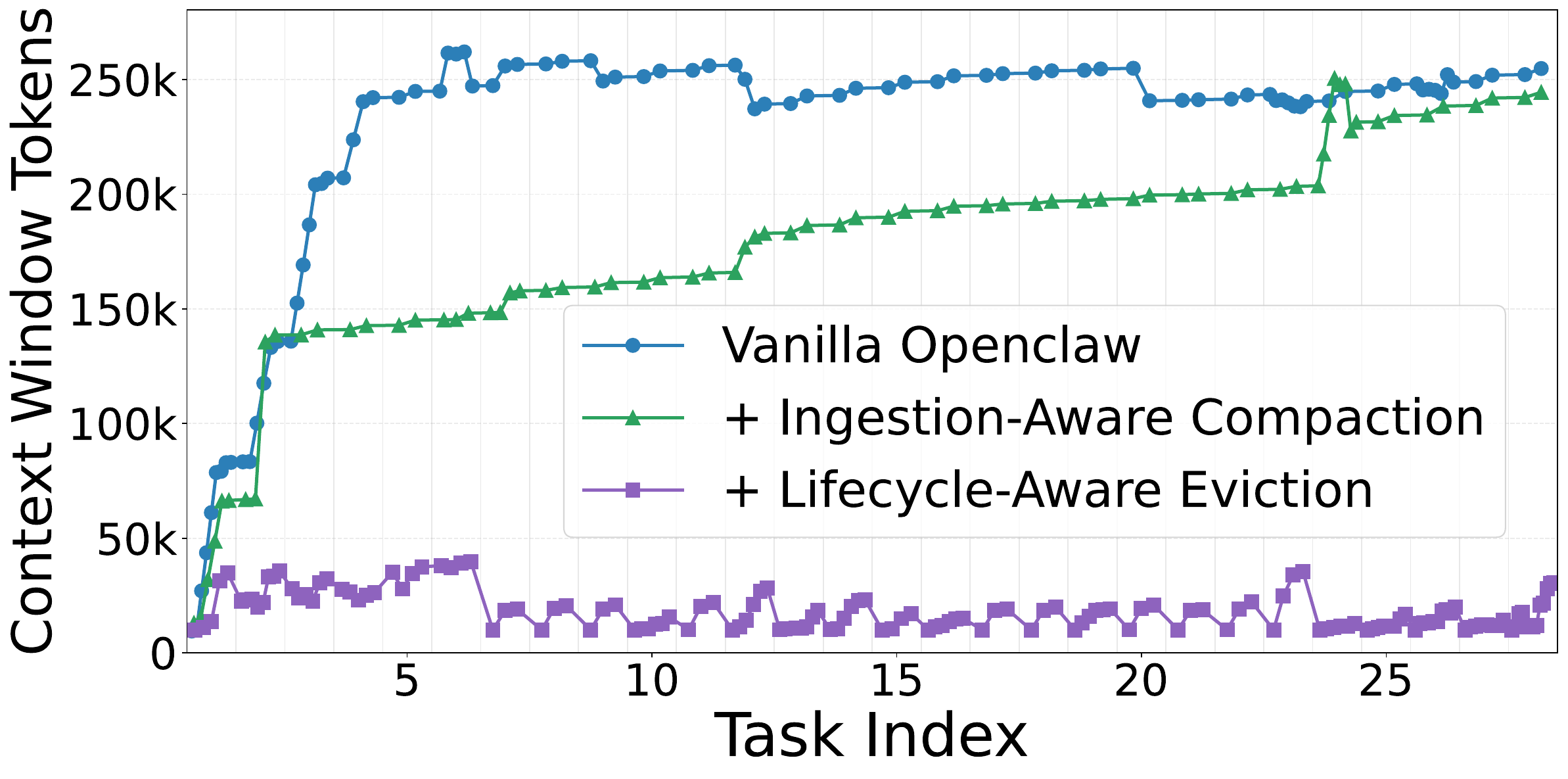}
    \caption{Per-call context token volume across a continuous 
    \texttt{Meeting Analysis} session.}
    \label{fig:eviction_trend}
\end{figure}
 
\subsection{Overall Performance}
 
Table~\ref{tab:pinchbench} and Table~\ref{tab:claw-eval} report the performance and resource consumption on \texttt{PinchBench} and \texttt{Claw-Eval} under both evaluation modes.

\paragraph{Isolated Mode.}
TokenPilot outperforms all evaluated baselines by securing the lowest total inference costs of \textbf{\$3.22} on \texttt{PinchBench} and \textbf{\$2.27} on \texttt{Claw-Eval} while maintaining competitive task accuracy. 
Text-level compression methods lower expenditures but consistently degrade task performance due to aggressive semantic pruning. 
Conversely, while dynamic frameworks preserve execution quality, they fail to regulate prompt cache and incur cache miss penalties. 
TokenPilot successfully bypasses these limitations, achieving optimal economy without sacrificing task effectiveness.

\paragraph{Continuous Mode.}
Under continuous task streams, long-horizon text accumulation severely amplifies these macro performance gaps. 
On \texttt{PinchBench}, TokenPilot sustains a top performance score of \textbf{81.3} at a minimal expenditure of \textbf{\$2.79}, restricting cache misses to \textbf{1.549M} tokens. 
On \texttt{Claw-Eval}, unrestricted history growth causes catastrophic cost inflation, forcing Vanilla expenditures to rocket to \textbf{\$81.52}. 
TokenPilot slashes this operational cost down to \textbf{\$10.58}, demonstrating robust scalability and systemic superiority over existing paradigms in deployment environments.

\subsection{Ablation Study}

Table~\ref{tab:ablation_progressive} and Figure~\ref{fig:eviction_trend} present the progressive contribution of each TokenPilot component against the Vanilla baseline, which lacks proactive entry-gate regulation. 
As shown in Figure~\ref{fig:eviction_trend}, the baseline context size climbs rapidly and remains persistently high across the execution horizon, despite its built-in compaction constraining the peak volume.

Integrating \textbf{Ingestion-Aware Compaction} via rule-based pruning and prefix stabilization slashes total expenditure from \textbf{\$7.24} to \textbf{\$4.22}.
This enhancement is validated by a sharp reduction in cache miss tokens from \textbf{5.943M} to \textbf{1.589M}. 
Explicitly, rule-based pruning dampens context peaks by filtering verbose environmental noise before insertion, while stabilization converts expensive pre-fills into cache hits by enforcing reliable layout reuse across consecutive tasks.

Layering \textit{Lifecycle-Aware Eviction} further minimizes expenditures to \textbf{\$2.79} while maintaining the overall score. 
This component triggers a \textbf{65.0\%} reduction in cache hit tokens from \textbf{26.716M} to \textbf{8.551M}, demonstrating that TokenPilot tightly caps the active memory footprint. 
The periodic tracking drops in Figure~\ref{fig:eviction_trend} confirm that our conservative batch-turn schedule executes precise memory offloading only when task residual utility expires.

\begin{figure*}[!t]
    \centering
    \subfigure[PinchBench Vanilla]{
        \label{fig:analysis_a} \includegraphics[width=0.49\linewidth]{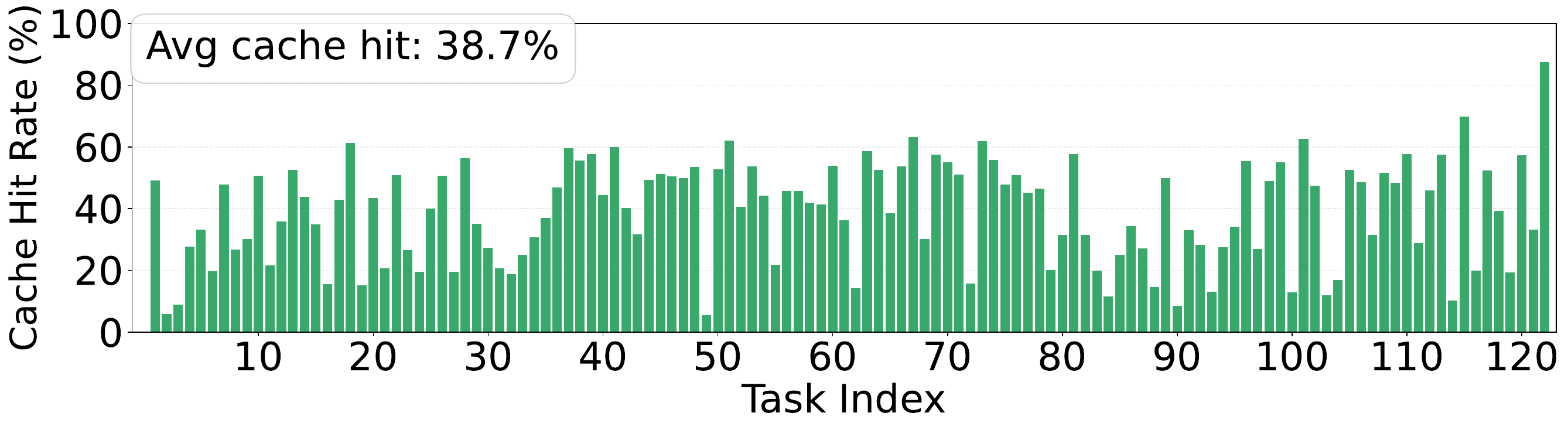}}
    \subfigure[PinchBench Stable Placeholders]{
        \label{fig:analysis_b} \includegraphics[width=0.49\linewidth]{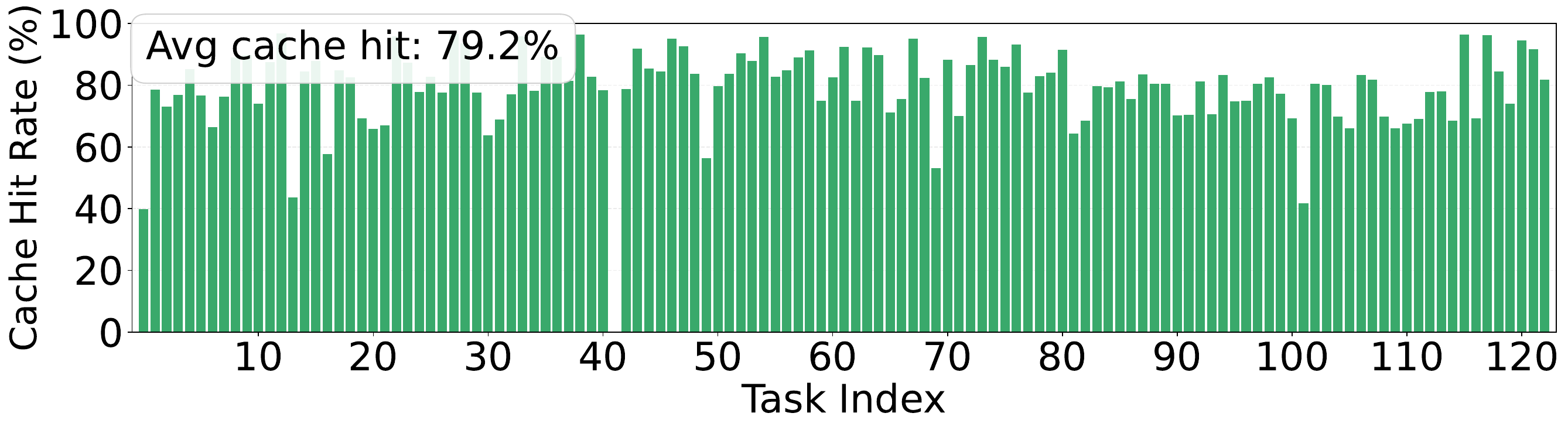}}
    \subfigure[Claw-Eval Vanilla]{
        \label{fig:analysis_c} \includegraphics[width=0.49\linewidth]{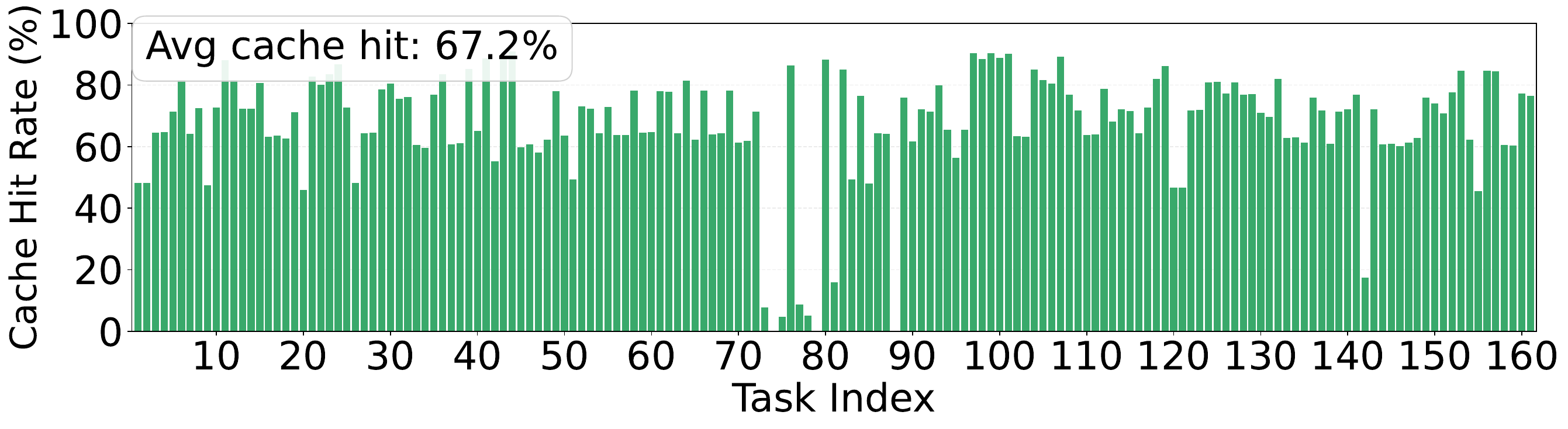}}
    \subfigure[Claw-Eval Stable Placeholders]{
        \label{fig:analysis_d} \includegraphics[width=0.49\linewidth]{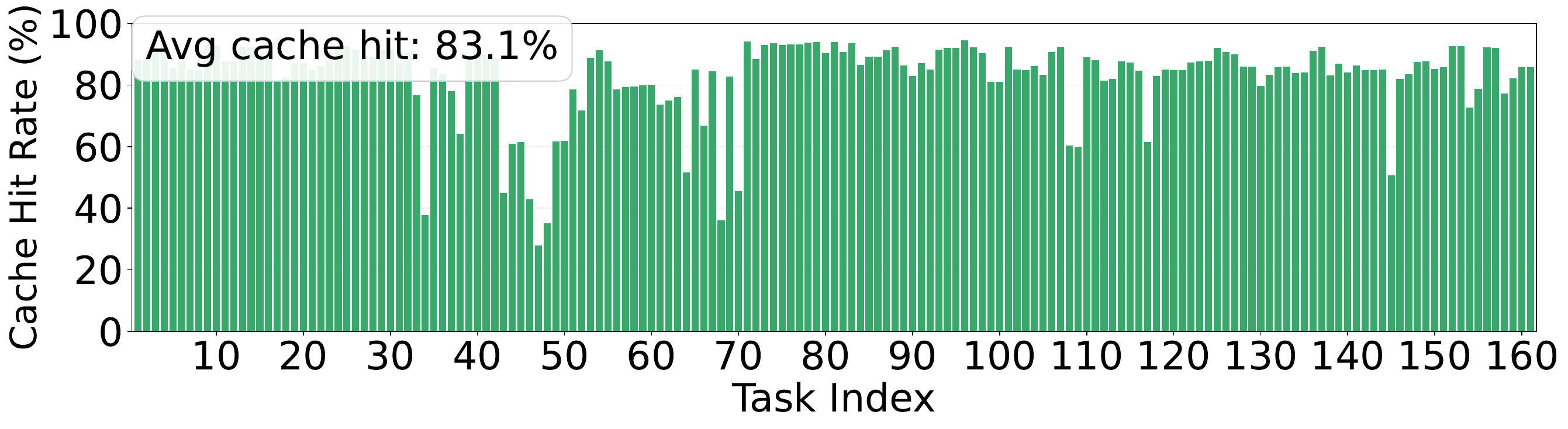}}
    \caption{Per-Task Cache Hit Rate on PinchBench and Claw-Eval.}
    \label{fig:Cache_stable}
\end{figure*}

\begin{figure*}[!t]
    \centering
    \subfigure[HTML Slimming Pass Reduction] { \label{fig:analysis_html} \includegraphics[width=0.49\linewidth]{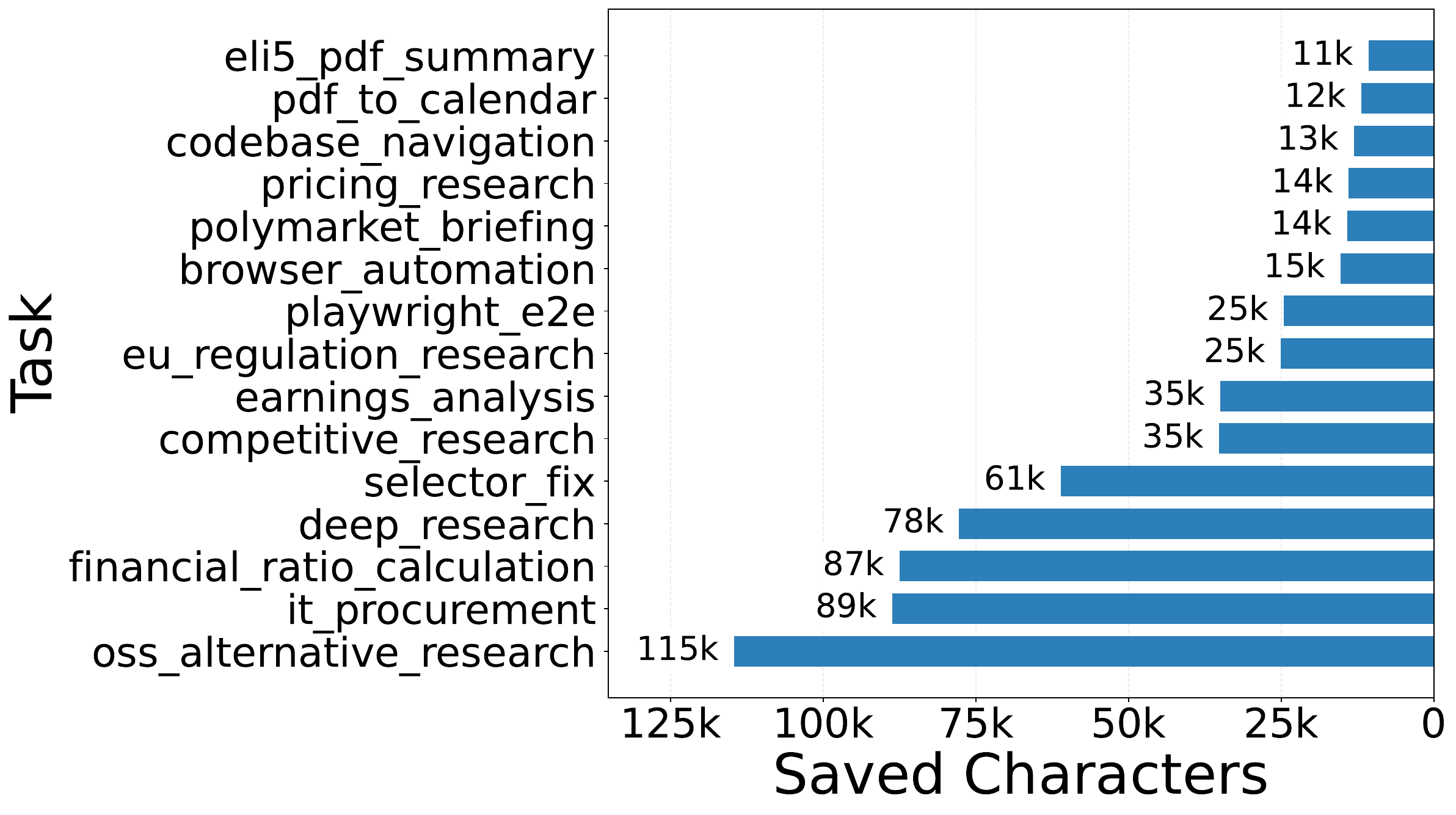}}
    \subfigure[Exec Output Truncation Pass Reduction] { \label{fig:analysis_exec} \includegraphics[width=0.49\linewidth]{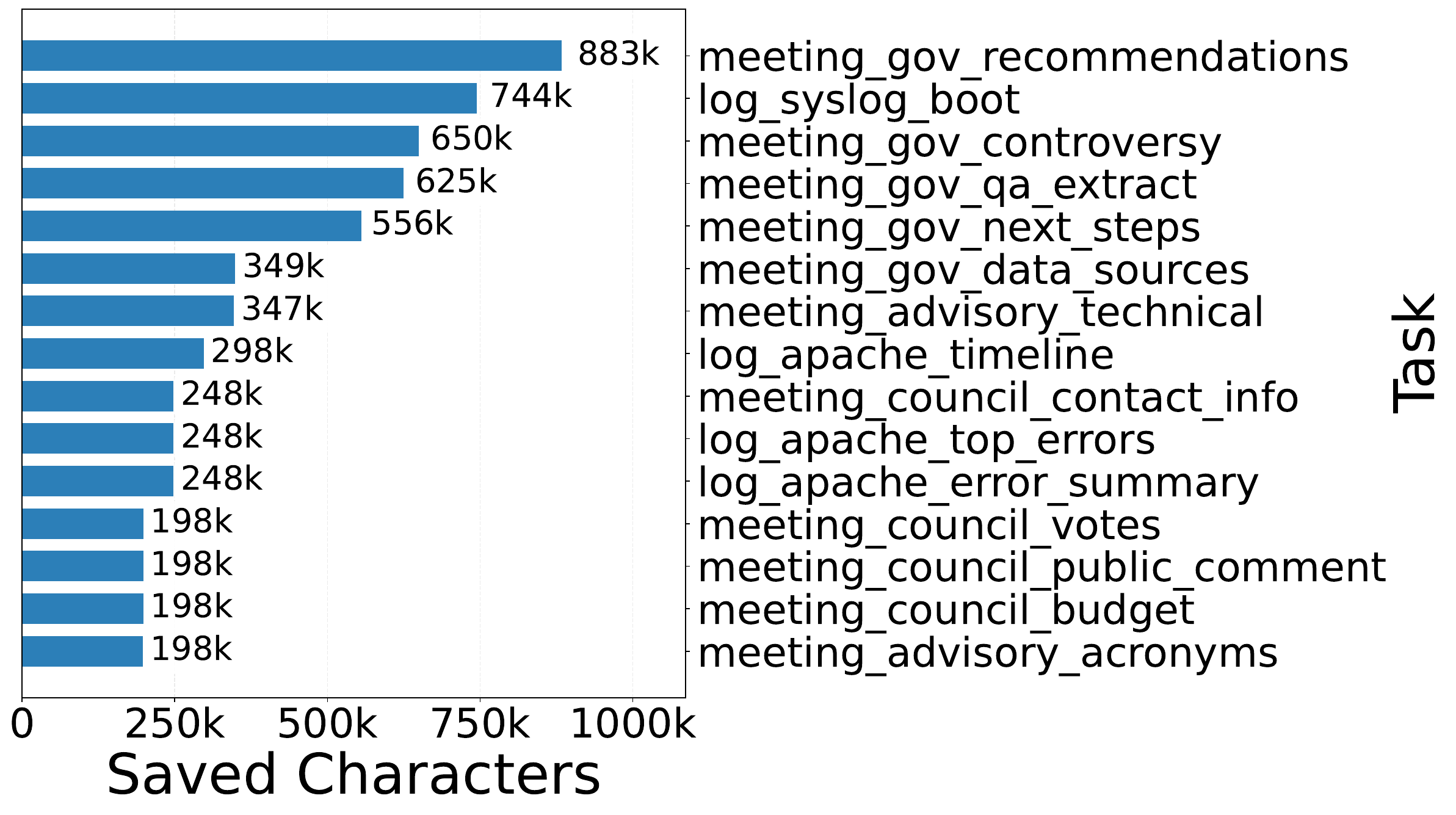}}
    \caption{Per-Task Character Savings from Reduction Passes.}
    \label{fig:Reduction}
\end{figure*}

\begin{table}[t]
\centering
{\setlength{\tabcolsep}{10pt}
\resizebox{\linewidth}{!}{%
\begin{tabular}{l|cc|ccc}
\toprule[1pt]
\textbf{Method} & \textbf{Overall} & \textbf{Cost (\$)} & \textbf{Hit (M)} & \textbf{Miss (M)} & \textbf{Output (M)} \\
\midrule
Vanilla & 80.47 & 8.31 & 6.184 & 8.753 & 0.285 \\
+ Cache Stabilization & 80.81 & 4.35 & 12.948 & 2.818 & 0.295 \\
+ Reduction Pass & \textbf{80.92} & \textbf{2.87} & 8.700 & 1.493 & 0.245 \\
- Recovery Tool & 77.12 & 4.03 & 11.780 & 2.539 & 0.276 \\
\bottomrule[1pt]
\end{tabular}
}}
\caption{Component-level analysis of \textit{Ingestion-Aware Compaction} 
on \texttt{PinchBench} in isolated mode.}
\label{tab:pinchbench_isolated}
\end{table}
\begin{table}[t]
\centering
\setlength{\tabcolsep}{8pt}
\resizebox{\linewidth}{!}{%
\begin{tabular}{c|cc|cc}
\toprule[1pt]
\multirow{2}{*}{\textbf{Cache Hit}} & \multicolumn{2}{c|}{\textbf{PinchBench}} & \multicolumn{2}{c}{\textbf{Claw-Eval}} \\
\cmidrule(lr){2-3} \cmidrule(lr){4-5}
& \textbf{Vanilla} & \textbf{w/ Stable} & \textbf{Vanilla} & \textbf{w/ Stable} \\
\midrule
0      & 8.1\%  & 2.44\%  & 100.0\% & 3.7\%  \\
2,048  & 91.9\% & 0.0\%  & 0.0\%   & 0.0\%  \\
5,120  & 0.0\%  & 77.24\% & 0.0\%   & 0.0\%  \\
5,888  & 0.0\%  & 0.0\%  & 0.0\%   & 0.6\%  \\
6,144  & 0.0\%  & 0.0\%  & 0.0\%   & 59.6\% \\
6,656  & 0.0\%  & 0.0\%  & 0.0\%   & 31.1\% \\
7,168  & 0.0\%  & 0.0\%  & 0.0\%   & 5.0\%  \\
12,288 & 0.0\%  & 6.50\%  & 0.0\%   & 0.0\%  \\
12,800 & 0.0\%  & 13.82\%  & 0.0\%   & 0.0\%  \\
\bottomrule[1pt]
\end{tabular}
}
\caption{Distribution of warm-start cache hit tokens at the 
first inference call of each task across benchmarks.}
\label{tab:first_call_cache}
\end{table}

\subsection{Analysis of Ingestion-Aware Compaction}
\label{sec:analysis_compaction}

We isolate the individual impacts of prefix stabilization and context reduction by benchmarking tasks independently in Table~\ref{tab:pinchbench_isolated}. Merely introducing stable placeholders cuts the baseline cost from \textbf{\$8.31} to \textbf{\$4.35} by converting cache misses into cache hits, while layering reduction passes further minimizes the expenditure to \textbf{\$2.87}.

\begin{table}[t]
\centering
\setlength{\tabcolsep}{4pt}
\resizebox{\linewidth}{!}{%
\begin{tabular}{l|cc|ccc}
\toprule[1pt]
\textbf{Method} & \textbf{Overall} & \textbf{Cost (\$)} & \textbf{Hit (M)} & \textbf{Miss (M)} & \textbf{Output (M)} \\
\midrule
Vanilla                         & 79.2 & 7.24 & 25.02 & 5.94 & 0.20 \\
\quad + Prefix Stabilization   & 79.4 & 6.39 & 28.02 & 4.40 & 0.22 \\
Summary                         & 78.4 & 7.12 & 20.69 & 6.25 & 0.20 \\
\quad + Prefix Stabilization   & 78.9 & 6.06 & 24.16 & 4.23 & 0.24 \\
Keep-Last-N                     & 79.1 & 5.66 & 18.12 & 4.48 & 0.21 \\
\quad + Prefix Stabilization   & 79.0 & 5.20 & 20.47 & 3.63 & 0.21 \\
\bottomrule[1.2pt]
\end{tabular}
}
\caption{Prefix stabilization applied to other baselines on \texttt{PinchBench} in continuous mode.}
\label{tab:prefix_stabilization_baselines}
\end{table}

\paragraph{Prefix Stabilization Facilitates Warm Starts.}
Physical cache continuity is disrupted by universal volatile fields, including directory paths and timestamps, and environment-specific tool definitions. Universal markers dominate prefix instability on \texttt{PinchBench}, whereas \texttt{Claw-Eval} configurations introduce severe tool-schema jitter. Replacing these dynamic fields with static placeholders transforms cold initializations into immediate warm starts, ensuring successive tasks inherit the accumulated prompt cache. 

As validated by Table~\ref{tab:first_call_cache}, stable placeholders migrate the vast majority of tasks from minimal baseline token allocations to high-capacity warm starts across both platforms. Consequently, Figure~\ref{fig:Cache_stable} shows that the macro cache hit rate rises from \textbf{38.7\%} to \textbf{79.2\%} on \texttt{PinchBench}, and from \textbf{67.2\%} to \textbf{83.1\%} on \texttt{Claw-Eval}.

Table~\ref{tab:prefix_stabilization_baselines} further applies stabilization to three representative context-management strategies under the same category-grouped continuous protocol. It reduces cost for Vanilla, Summary, and Keep-Last-N. The smaller gain for Keep-Last-N reflects a remaining limitation of fixed-window truncation: stabilization preserves the invariant core prefix, but cannot recover task-specific cache reuse after the retained history window shifts.

\paragraph{Context Reduction Compounds Savings via Fallback Loops.}
Context reduction exploits a compounding logic: tokens eliminated at the framework boundary never accumulate in multi-turn windows. Figure~\ref{fig:Reduction} demonstrates that our dual reduction passes directly strip heavy payloads across heterogeneous tasks, removing up to \textbf{115k} characters of structural noise in \texttt{oss\_alternative\_research} via HTML slimming, and up to \textbf{883k} characters of terminal logs in \texttt{meeting\_gov\_recommendations} via execution truncation. This targeted compaction effectively trims the sequence footprint while sustaining task accuracy at \textbf{80.9}.

The recovery tool is essential to sustaining this performance boundary. Completely disabling it triggers a capability drop from \textbf{80.9} to \textbf{77.1} while inflating expenditures to \textbf{\$4.03}. Without full content access, the agent executes compensatory retries that append fresh tool feedback and overwhelm rule-based compaction. The recovery mechanism breaks this inflationary cycle by providing on-demand payload access, preserving task effectiveness while halting uncontrolled context growth.
 
\subsection{Analysis of Lifecycle-Aware Eviction}

\begin{figure}[t]
    \centering
    \includegraphics[width=\linewidth]{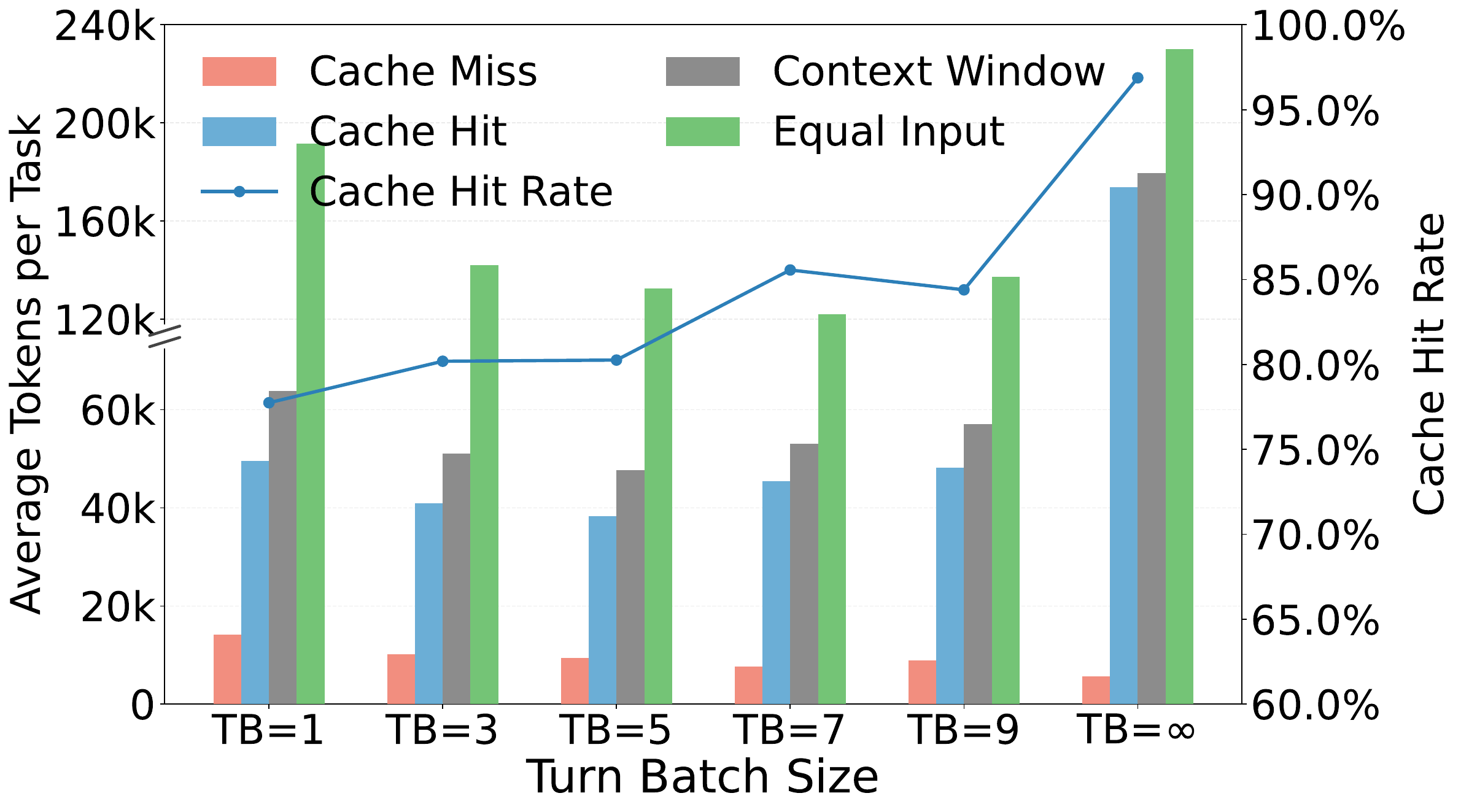}
    \caption{Average per-task Cache Miss, Cache Hit, Context Window, and Equal Input tokens, alongside Cache Hit Rate, across turn batch sizes $B \in \{1, 3, 5, 7, 9, \infty\}$ on the \texttt{Meeting Analysis} category of \texttt{PinchBench} in continuous mode.}
    \label{fig:tb_analysis}
\end{figure}

\paragraph{Batch Trigger Intervals Regulate Context Size and Cache Stability.}
We evaluate eviction trigger frequencies in Figure~\ref{fig:tb_analysis}, where \textit{Context Window} tracks physical text accumulation, and \textit{Equal Input} measures the equivalent monetary cost by discounting cache hits relative to expensive pre-fill misses. 
The \textit{Cache Hit Rate} line reflects the percentage of cached tokens relative to total ingestion.

Completely disabling eviction ($B = \infty$) causes both metrics to peak, confirming that unbounded history growth escalates deployment expenditures. While lifecycle eviction lowers both curves, a hyperactive schedule ($B = 1$) triggers premature truncation that disrupts layout consistency and inflates cache misses. Conversely, larger batch sizes preserve prefix continuity to improve cache hit rates. Balancing task accuracy, memory reduction, and API call times, $B = 3$ constitutes the empirical optimum by preventing memory inflation while securing reliable hardware-level cache reuse.

\begin{table}[t]
\centering
\setlength{\tabcolsep}{4pt}
\resizebox{\linewidth}{!}{%
\begin{tabular}{l|cc|ccc}
\toprule[1pt]
\textbf{Retention Signal} & \textbf{Overall} & \textbf{Cost (\$)} & \textbf{Hit (M)} & \textbf{Miss (M)} & \textbf{Output (M)} \\
\midrule
\multicolumn{6}{l}{\textit{Category-grouped streams}} \\
Recent-turn             & 78.97 & 5.86 & 12.94 & 5.00 & 0.25 \\
Recent-file-access      & 74.11 & 5.17 & 14.49 & 3.98 & 0.24 \\
Tool-dependency         & 78.06 & 6.82 & 20.85 & 5.63 & 0.23 \\
Task-category           & 78.71 & 5.31 & 19.71 & 3.66 & 0.24 \\
Residual-utility        & 78.81 & 4.69 & 17.16 & 2.99 & 0.26 \\
\midrule
\multicolumn{6}{l}{\textit{Mixed-category streams}} \\
Recent-turn             & 69.80 & 5.79 & 18.34 & 4.31 & 0.26 \\
Recent-file-access      & 67.60 & 6.38 & 23.95 & 4.38 & 0.29 \\
Tool-dependency         & 67.71 & 7.04 & 25.54 & 5.14 & 0.28 \\
Task-category           & 67.59 & 6.77 & 23.72 & 5.03 & 0.27 \\
Residual-utility        & 69.16 & 5.25 & 14.83 & 3.85 & 0.28 \\
\bottomrule[1.2pt]
\end{tabular}
}
\caption{Controlled comparison of lifecycle retention signals on \texttt{PinchBench} in category-grouped and mixed-category continuous streams. Observation reduction is disabled, and all methods share prefix stabilization and the same block-level eviction executor.}
\label{tab:lifecycle_retention_signals}
\end{table}

\begin{figure}[t]
    \centering
    \includegraphics[width=\linewidth]{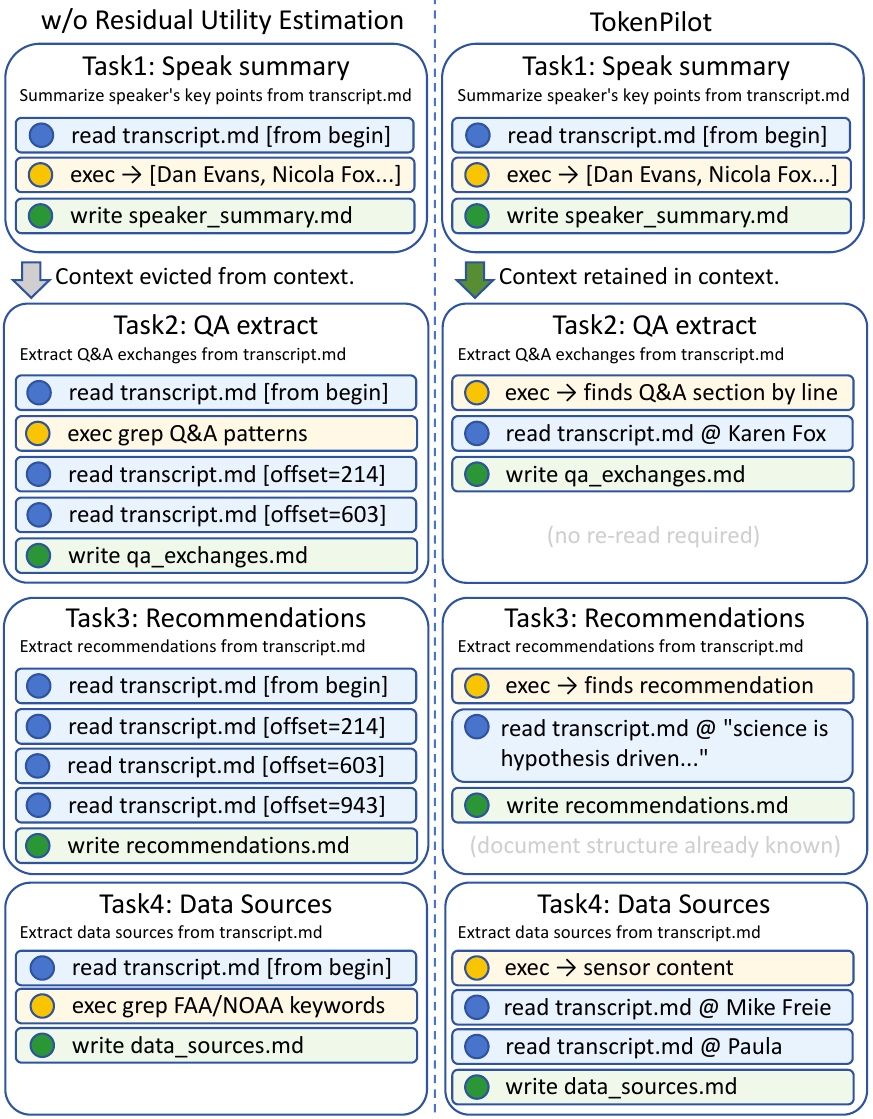}
    \caption{Tool call patterns across a four-task session on \texttt{transcript.md} under TokenPilot and a variant without residual utility estimation. TokenPilot retains the context segment after task completion, enabling subsequent tasks to directly access relevant document sections. Without residual utility estimation, each task re-reads files from the beginning and issues multiple sequential partial reads to locate the same content.}
    \label{fig:case_study}
    \end{figure} 
 
\paragraph{Retention Signal Comparison.}

To isolate the retention decision from the eviction implementation, Table~\ref{tab:lifecycle_retention_signals} compares our model-based residual-utility estimator with four interpretable signals. Recent-turn retains tasks active within the last three turns; recent-file-access retains tasks associated with a file read within the last three turns and not subsequently overwritten; tool-dependency retains tasks sharing a tool family with the last three turns; and the task-category retains the current contiguous category segment. All variants use the same runtime, backbone, continuous protocol, prefix stabilization, and block-level eviction executor, with observation reduction disabled.

Because grouping same-category tasks into contiguous sessions can favor cross-task residual utility and cache reuse, we additionally evaluate heterogeneous streams constructed by shuffling tasks across categories (see Appendix~\ref{appendix:mixed} for detailed setting). On the category-grouped streams, the residual-utility estimator achieves the lowest cost of \$4.69 while retaining a score comparable to the strongest heuristic. On mixed-category streams, it remains the lowest-cost signal and reduces cost to \$5.25 relative to Recent-turn. The narrower margin is expected because frequent task and tool-schema changes leave less residual utility to exploit. Since all variants share the same execution path, these differences arise from the retention signal rather than from different deletion or rewrite mechanisms. 

\paragraph{Residual Utility Prevents Over-Eviction.}
\label{Residual}

We complement this aggregate comparison with a variant that disables residual utility estimation and immediately purges historical segments upon sub-task completion. Figure~\ref{fig:case_study} traces a representative four-task session in which successive tasks manipulate a shared file named \texttt{transcript.md}.
   
Under TokenPilot, the primary task ingests \texttt{transcript.md} and anchors its structure in memory. 
When downstream tasks arrive, the framework estimator infers from tool dependency patterns that this historical segment retains residual utility, preserving its slots despite sub-task completion. 
Consequently, tools directly locate target content blocks including personnel sections and roadmap data without re-exploring background knowledge.

Conversely, the variant without residual utility estimation triggers eviction immediately upon local execution completion. 
Each subsequent task thus inherits a cold window, forcing it to rediscover the document architecture via redundant full file reads and sequential scans. 
This contrast demonstrates that the residual utility buffer acts as a valve to preserve document knowledge across distinct tasks, enabling targeted access while blocking the overhead of continuous context re-exploration.

\section{Related Work} \label{Relatedwork}

\paragraph{Static Content Compression and Abstraction.} 
From the perspective of maximizing utility density within the context window, a prominent strategy focuses on filtering or restructuring historical trajectories at the ingestion stage to preserve the premium context budget. 
To eliminate linguistic redundancy and limit token footprints, early efforts successfully prune non-essential text units or compress structural prompt representations~\cite{llmlingua, llmlingua2, SelectiveContext, swezze, TokenReduction}. 
Extending this content-reduction philosophy to a macro scale, passive memory retrieval systems manage the runtime working memory by offloading entire session histories to external databases, selectively recalling high-utility historical fragments while filtering out the remaining interactions~\cite{packer2023memgpt, chhikara2025mem0, memorybank, lightmem}. 
Rather than applying hard textual pruning, complementary approaches enhance information density by converting raw trajectories into structured, high-level semantic abstractions. 
To maintain holistic task coherence, these frameworks effectively substitute continuous interaction footprints with localized recursive summaries, hierarchical sub-goal graphs, or episodic, temporal knowledge graphs to secure macro-level guidance across elongated session horizons~\cite{lcm2026, resum, hiagent, CAT, DeepAgent, rasmussen2025zep}.

\paragraph{Dynamic and Runtime Scheduling.} 
Another significant line of research treats the context window as a fluid operating system resource, managing context segments in real time to align with the agent's immediate execution states. 
To optimize token distribution during live sessions, modern frameworks successfully execute runtime demand paging, adaptive parallel routing, and context isolation based on real-time trajectories and budget constraints~\cite{Pichay, agentswing, memobrain, ContextBudget, agentfold, ContextFolding, LatentContextManager}. 
For long-horizon planning, advanced architectures manage memory as a self-organizing operating system or introduce virtual memory abstractions to dynamically secure stateful data residency and tool durability~\cite{structmem, mobilemem,lightmem-ego, FluxMem, MemOS, MemoryOS, ClawVM}. 
Recently, this runtime scheduling paradigm has expanded to multi-agent environments, utilizing decentralized role-aware routing, centralized experience caching, or cross-context KV-cache communication topographies to minimize distributed token footprints across collaborative swarms~\cite{RCRRouter, stackplanner, legomem, KVCOMM}.
\section{Conclusion}  \label{Conclusion}
We presented \textbf{TokenPilot}, a dual-granularity framework that reconciles text reduction with strict prompt cache alignment. 
By separating memory management into global ingestion-aware compaction and local lifecycle-aware eviction, our framework successfully stabilizes dynamic layouts while conservatively offloading context based on task-level residual utility. 
Empirical evaluations on both \texttt{PinchBench} and \texttt{Claw-Eval} demonstrate that \textbf{TokenPilot} drastically cuts inference expenditures under both isolated and continuous operational modes without sacrificing task effectiveness, offering a scalable and highly cost-efficient foundation for long-horizon agent systems.
\section*{Limitations}
Despite its strong performance, TokenPilot has several limitations. 
The model-based estimator may misclassify context segments under highly ambiguous or sparse interaction patterns, and the frequency threshold $\tau$ and batch size $B$ may require tuning for different deployment environments and task distributions. 
The prefix stabilization component additionally relies on backend support for prefix caching, providing no benefit to providers without this feature. 
Finally, our continuous mode evaluation groups same-category tasks into contiguous sessions to reflect domain-specific workflow environments; heavily shuffled or highly heterogeneous mixed-category task streams may naturally exhibit lower prefix reuse rates due to persistent tool schema mutations, which we leave as an important direction for future investigation.

\section*{Acknowledgements}
We would like to express our sincere gratitude to the anonymous reviewers for their thoughtful and constructive feedback. This work was supported by the New Generation Artificial Intelligence-National Science and Technology Major Project (2025ZD0122802), National Natural Science Foundation of China (No. 62576307, No. NSFCU23B2055, No. NSFCU19B2027), the Fundamental Research Funds for the Central Universities (226-2023-00138), the Yongjiang Talent Introduction Programme (2021A-156-G), and the Information Technology Center and State Key Lab of CAD\&CG, Zhejiang University.  

\bibliography{custom}

\appendix
\section{Appendix}

\subsection{Dataset Configurations}
\label{Appendix:Dataset}

To evaluate \texttt{TokenPilot}, we utilize two realistic agent benchmarks: \texttt{PinchBench} and \texttt{Claw-Eval}. 

\texttt{PinchBench} is a real-world evaluation suite comprising 11 distinct task categories and 123 tasks in total. 
To account for the continuous rolling updates in the upstream repository, we benchmark our framework on a frozen snapshot of the benchmark.\footnote{\url{https://github.com/pinchbench/skill/commit/0347a7f1736a9c33b5fe831e27d1d6ee9b576221}}

\texttt{Claw-Eval} is a containerized agent evaluation platform executed within isolated sandbox environments. 
We evaluate on its \textbf{General} task group, which encompasses 161 multi-step service orchestration and standalone analytical tasks. 

For both benchmarks, we group same-category tasks into contiguous, uninterrupted single sessions to faithfully simulate realistic continuous multi-task agent execution trajectories. 
The detailed structural statistics of these evaluation platforms are compiled in Table~\ref{tab:dataset_stats}.

\begin{table}[t]
\centering
\resizebox{\linewidth}{!}{%
\begin{tabular}{l|c||l|c}
\toprule[1pt]
\multicolumn{2}{c||}{\textbf{PinchBench}} & \multicolumn{2}{c}{\textbf{Claw-Eval (General)}} \\
\cmidrule(lr){1-2} \cmidrule(lr){3-4}
\textbf{Category} & \textbf{\#} & \textbf{Category} & \textbf{\#} \\
\midrule
Productivity     & 8  & Workflow       & 47 \\
Research         & 12 & Ops            & 31 \\
Writing          & 6  & Finance        & 14 \\
Coding           & 14 & Office QA      & 10 \\
Analysis         & 12 & Communication  & 8  \\
CSV Analysis     & 26 & Productivity   & 7  \\
Log Analysis     & 6  & Operations     & 6  \\
Meeting Analysis & 28 & Safety         & 5  \\
Memory           & 2  & Terminal       & 5  \\
Skills           & 6  & Multimodal     & 4  \\
Integrations     & 3  & Others         & 24 \\
\midrule
\textbf{Total}   & \textbf{123} & \textbf{Total} & \textbf{161} \\
\bottomrule[1pt]
\end{tabular}
}
\caption{Statistics for \texttt{PinchBench} and \texttt{Claw-Eval}.}
\label{tab:dataset_stats}
\end{table}

\subsection{Evaluation Metrics and Cost Modeling}
\label{Appendix:Metrics}

\paragraph{Task Score Execution Framework.}
We strictly adhere to the native evaluation frameworks provided by each respective benchmark to compute task performance. 

For \texttt{Claw-Eval}, the scoring pipeline executes an integrated multi-dimensional protocol evaluating \textit{Completion} ($s_{\text{comp}}$), \textit{Safety} ($s_{\text{safe}}$), and \textit{Robustness} ($s_{\text{rob}}$) as coupled parameters grounded in multi-channel auditable trajectory evidence, including service audit logs, environment snapshots, and execution traces. 
Formally, the final task score is mathematically formulated as follows:
\begin{equation}
    \text{Score} = s_{\text{safe}} \times (0.80 \cdot s_{\text{comp}} + 0.20 \cdot s_{\text{rob}})
\end{equation}
where Safety acts as a strict multiplicative gate, while Completion and Robustness represent the primary goal-directed execution quality and secondary error-recovery capability under controlled service perturbations respectively. 
This fine-grained rubric triangulation yields continuous partial credits rather than trivial binary verdicts. 

For \texttt{PinchBench}, the framework aggregates task-specific verification checks on output deliverables to assess the agent's goal-directed capability. 

Crucially, when evaluating system trajectories in \textbf{Continuous Mode}, we implement a trajectory slicing mechanism that automatically partitions the continuous session transcript file into task-specific segments based on original task boundaries. 
Each sliced segment is then fed independently into the corresponding benchmark grader, ensuring that the evaluation logic for continuous task streams remains strictly identical and mathematically comparable to that of the \textbf{Isolated Mode}.

\paragraph{Inference Cost Modeling.}
To calculate the runtime inference cost across sequential execution sessions, we implement a monetary cost metric based on commercial deployment pricing. 
The total inference cost is calculated as follows:
\begin{equation}
    \text{Cost} = |\mathcal{C}'_{\text{hit}}| \cdot p_{\text{hit}} + |\mathcal{C}'_{\text{miss}}| \cdot p_{\text{miss}} + \mathcal{H}_{\text{out}} \cdot p_{\text{out}}
\end{equation}
where $|\mathcal{C}'_{\text{hit}}|$ and $|\mathcal{C}'_{\text{miss}}|$ represent the number of input tokens that hit or miss the KV cache backend respectively, and $\mathcal{H}_{\text{out}}$ represents the length of the generated agent responses.
Following the official pricing tiers of \texttt{GPT-5.4-mini}, the price parameters are set to $p_{\text{hit}} = \$0.075/\text{M}$ tokens for cache hits, $p_{\text{miss}} = \$0.75/\text{M}$ tokens for cache misses, and $p_{\text{out}} = \$4.50/\text{M}$ tokens for outputs.

\subsection{Baseline Configurations} \label{appendix:baseline}
To ensure reproducibility, we document the configurations for all evaluated baselines.

\textcircled{\scriptsize 1} \textbf{Vanilla} runs on OpenClaw without any extra context management, with a maximum context window of 500k tokens and a compaction trigger ratio of 0.5.

\textcircled{\scriptsize 2} \textbf{LLMLingua-2} applies token-level compression using a small language model, with a compression ratio of 0.6.

\textcircled{\scriptsize 3} \textbf{SelectiveContext} applies sentence-level compression based on self-information, with a compression ratio of 0.4.

\textcircled{\scriptsize 4} \textbf{LCM} applies lossless compaction via hierarchical summarization, triggered when context reaches 75\% of the context window. Each leaf chunk accumulates up to 80k tokens before summarization, retaining the 64 most recent turns in full fidelity.

\textcircled{\scriptsize 5} \textbf{Pichay} uses utility-driven demand paging with the following thresholds: advisory zone at 60k tokens, involuntary eviction zone at 100k tokens, and a hard cap at 120k tokens. Tool results older than 4 user turns are eligible for compression, with a minimum eviction size of 500 bytes.

\textcircled{\scriptsize 6} \textbf{Summary} compresses interaction history into hierarchical summaries when context reaches 40\% of the 500k token window.

\textcircled{\scriptsize 7} \textbf{MemoBrain} maintains a memory budget of 100k tokens and triggers recall when estimated context length reaches 35\% of the memory budget.

\textcircled{\scriptsize 8} \textbf{AgentSwing} selects among three candidate strategies (discard-all, keep-last-n, summary) via lookahead simulation of 3 future turns. It triggers at a token ratio of 0.2 within a 200k context window, retaining the 5 most recent turns under the keep-last-n strategy.

\textcircled{\scriptsize 9} \textbf{Keep-Last-N} retains the most recent $N=5$ turns when context reaches 40\% of the 500k token window.

\textcircled{\scriptsize 10} \textbf{MemOS} limits retrieval to 20 items per recall turn to control token costs. It filters candidate memories by exposing at most 500 characters per item to the validation model, while restricting individual memory ingestion to a maximum of 20,000 characters per message.

\subsection{Implementation Details}
\label{appendix:Implementation}

This section documents the underlying engineering configurations, threshold parameterizations, and prompting architectures of \texttt{TokenPilot} to facilitate exact reproducibility. 
We detail the deterministic mechanics for cache stabilization and observation reduction in alignment with our system design, followed by their fine-grained numerical hyperparameters and specific base model assignments. 
Finally, we present the complete system prompt templates utilized for our state estimation pipeline.

\paragraph{Cache Stabilization.}
To ensure the prompt prefix remains byte-identical across consecutive turns, \texttt{TokenPilot} standardizes and restructures the input layout before each inference call. 

First, runtime-volatile text fields within the system prompt messages, such as working directory paths, active timestamps, and transient session identifiers, are substituted with static, stable placeholders. 
Second, since distinct tasks often require different tool configurations, leaving tool definitions inside the primary system prompt introduces structural variations that break baseline prefix matching. 
To mitigate this positioning jitter, we systematically relocate the tool definitions and schemas downstream, placing them at the end of the system prompt message directly alongside the dynamic context block containing the original values of the volatile fields.

\paragraph{Observation Reduction.}
To suppress textual redundancy and regulate the per-turn input volume, we implement a sequence of rule-based reduction passes targeting low-utility tool result messages before they enter the canonical history. 
Specifically, repeated tool call results are deduplicated via hashing, while oversized tool call parameters and long execution outputs are truncated beyond a fixed token threshold. 
To prevent critical information loss from hard truncation, we equip the agent with a dedicated recovery tool, allowing it to dynamically retrieve full execution payloads when necessary.
For multimodal and web-browsing interactions, web-fetched content undergoes HTML slimming to remove non-essential markup and attributes, and embedded images are downsampled to minimize their respective token footprint. 
Finally, a general formatting pass cleans up the remaining layout variations by removing code fences, invalid format symbols, and line number prefixes from code outputs, alongside normalizing continuous whitespace characters.

\paragraph{Hyperparameters for Context Reduction.}
\label{paragraph:hyperparameters}
For the rule-based context reduction passes, the specific numerical thresholds and fine-grained hyperparameter configurations are parameterized as follows:

\textcircled{\scriptsize 1} \textbf{Activation Gates:} The minimum character count to trigger before-call reduction is set to \texttt{triggerMinChars = 2200}, and candidate tool-like fragments are routed to the module only when exceeding \texttt{maxToolChars = 1200}.

\textcircled{\scriptsize 2} \textbf{Execution Output Truncation:} The global truncation threshold for generic tool feedback is bounded at \texttt{50k} characters. For tool-specific profiles, the limits are set to \texttt{30k} for \texttt{bash/shell/powershell}, \texttt{20k} for \texttt{grep/rg}, \texttt{10k} for \texttt{mcp\_auth}, and \texttt{100k} for \texttt{glob/write/edit}, while \texttt{read/file\_read} is permitted an unconstrained capacity (\texttt{Infinity}). Truncated outputs consistently retain an initial \texttt{600}-character prefix and a terminal \texttt{400}-character suffix as a preview block, which can be fully recalled via the recovery tool when triggered by the agent.

\textcircled{\scriptsize 3} \textbf{Deduplication and Frequency Limits:} For the \texttt{repeated\_read\_dedup} pass, redundant read operations are substituted with the same \texttt{600/400} preview block. To suppress infinite loop behaviors and redundant footprint accumulation, the maximum sequential execution frequency for any identical tool call within a rolling tracking window is strictly capped at \texttt{5}. Multimodal constraints under \texttt{image\_downsample} restrict standard bitmap layouts to a maximum size of \texttt{100KB} and vector-based \texttt{SVG} documents to \texttt{50KB}.

\textcircled{\scriptsize 4} \textbf{Layout Cleaning Constraints:} File path markers are clipped via \texttt{path\_truncation} to a maximum length constraint of \texttt{80} characters. The remaining syntactic layout transformations, including \texttt{html\_slimming}, \texttt{format\_slimming}, \texttt{format\_cleaning}, and \texttt{line\_number\_strip}, are deterministically executed without extra numerical hyperparameters.

\paragraph{Model and Hyperparameter Configurations.}
\label{paragraph:models}
To ensure a rigorous and fair empirical comparison, all baseline agent architectures and our primary inference execution module are deployed under identical base model configurations, specifically utilizing \texttt{GPT-5.4-mini}. For the internal state estimation and context utility metrics, we employ \texttt{Qwen3.5-35B-A3B} as the dedicated estimator model. The batch-turn tracking window for interval context processing is set to $3$. 

\paragraph{Prompt Templates for the State Estimator} 
\label{appendix:prompts}
\begin{figure*}[t]
    \centering
    \includegraphics[width=0.9\textwidth]{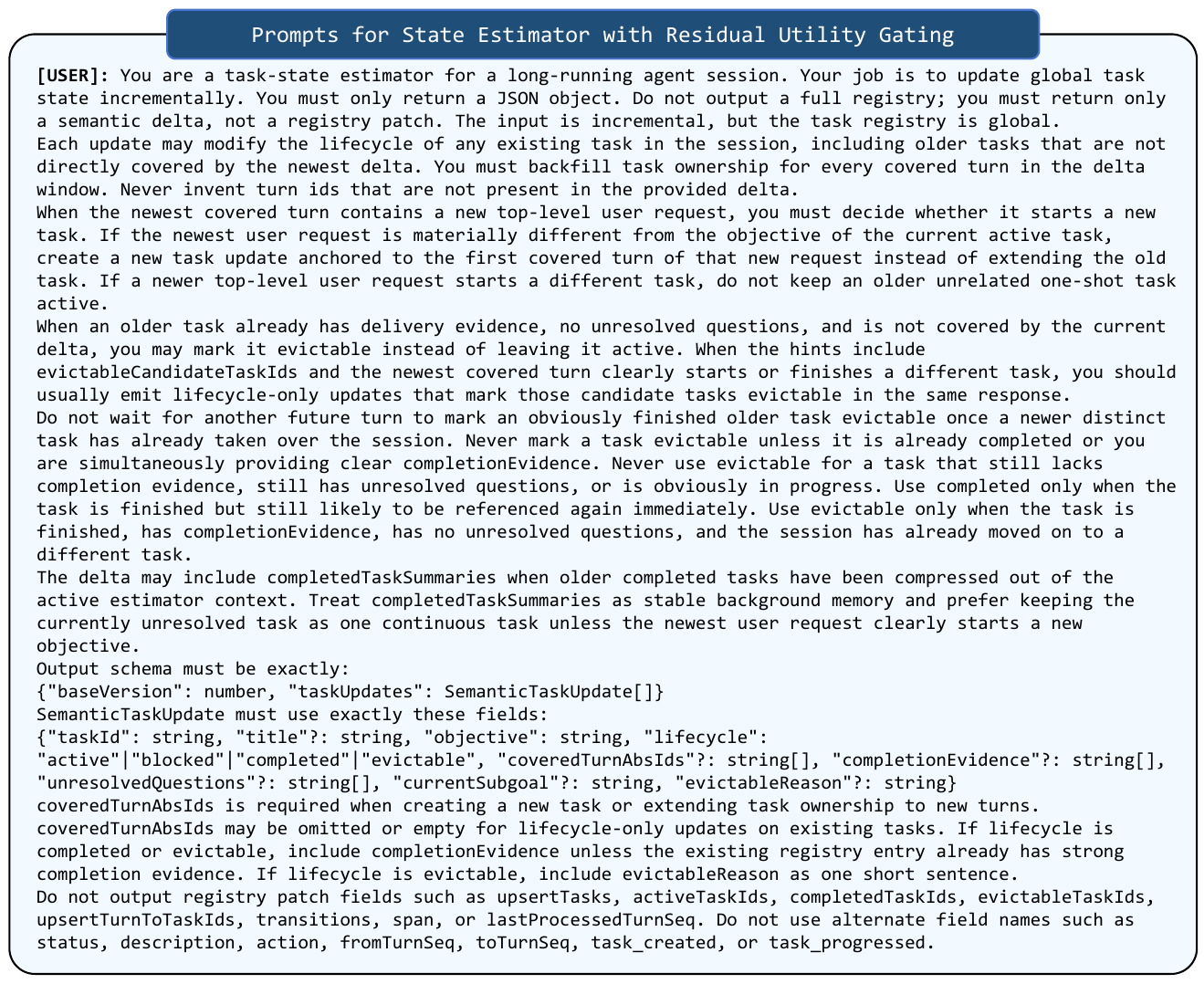}
    \caption{System prompt template for TokenPilot's Primary Estimator, featuring joint tracking of completion evidence and explicit cache eviction signaling.}
    \label{fig:prompt_1}
\end{figure*}

\begin{figure*}[t]
    \centering
    \includegraphics[width=0.9\textwidth]{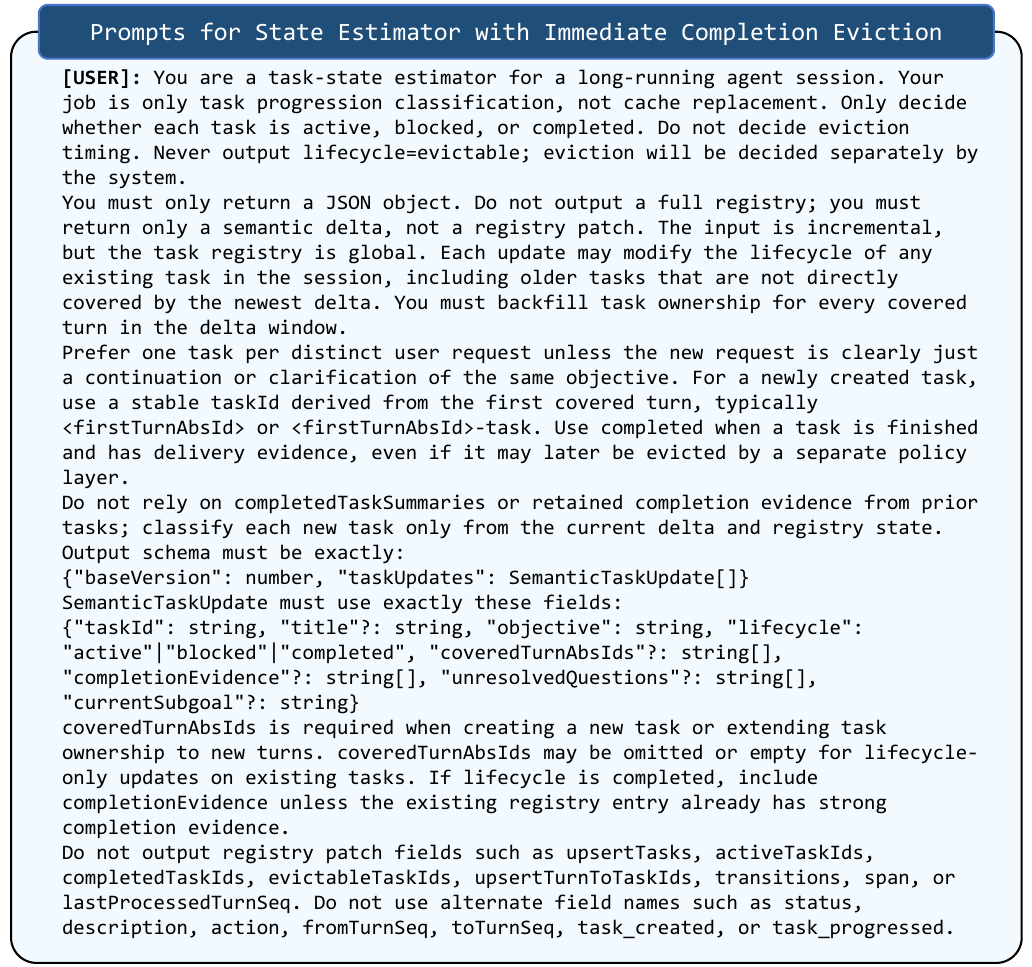}
    \caption{System prompt template for the Estimator without Residual Utility Estimation, configured to strip out the caching buffer for the ablation study.}
    \label{fig:prompt_2}
\end{figure*}
To provide full architectural transparency and ensure exact reproducibility, we detail the core system prompt configurations injected into our \texttt{Qwen3.5-35B-A3B} state estimator. 
The estimator operates as a structured semantic tracking pipeline that processes continuous session trajectories and outputs incremental semantic deltas in a validation-ready JSON format. 
Depending on whether the intermediate residual utility gating layer is active, the system configures the estimator prompt under two distinct tracking paradigms:

As illustrated in Figure~\ref{fig:prompt_1}, the full operational configuration of \texttt{TokenPilot} deploys a joint classification-and-eviction scheme. 
The system prompt instructs the estimator to evaluate ongoing tool dependencies and cross-turn data reuse patterns before rendering an expiration judgment. 
Crucially, it introduces a three-state transition matrix by enforcing an explicit \texttt{evictable} lifecycle token alongside standard \texttt{active} and \texttt{completed} identifiers.
By verifying delivery evidence and historical dependencies across the session trajectory, this prompt establishes a text-level buffer gate. 
Historical context segments are only flagged for cache clearance when the estimator explicitly infers that their operational task relevance has fully expired.

To systematically isolate the impact of our gating mechanism, the ablated configuration documented in Figure~\ref{fig:prompt_2} strips out the cache-aware buffering layer. 
Under this setup, the prompt restricts the model's objective to binary task progression classification. 
It limits the lifecycle field to \texttt{active} or \texttt{completed} tokens and prohibits the output of the \texttt{evictable} status identifier. 
Tasks transition directly to \texttt{completed} when local delivery evidence is observed, which triggers immediate context purging at the hardware backend. 
This configuration serves as the baseline to evaluate the precise cost and efficiency gains brought by the residual utility inference mechanism discussed in Section~\ref{Residual}.

\subsection{Mixed-Category Continuous Evaluation}
\label{appendix:mixed}

The continuous evaluation in the main experiments groups tasks from the same category into contiguous sessions, modeling domain-focused workflows. To test a more heterogeneous boundary condition, we additionally construct mixed-category streams by shuffling all 123 \texttt{PinchBench} tasks with fixed seed 42 and distributing them into ten continuous sessions of 12 or 13 tasks each. This protocol introduces frequent task-category and tool-schema changes and therefore reduces both cross-task residual utility and opportunities for prefix reuse.

\begin{table}[t]
\centering
\setlength{\tabcolsep}{4pt}
\resizebox{\linewidth}{!}{%
\begin{tabular}{l|cc|ccc}
\toprule[1pt]
\textbf{Method} & \textbf{Overall} & \textbf{Cost (\$)} & \textbf{Hit (M)} & \textbf{Miss (M)} & \textbf{Output (M)} \\
\midrule
Vanilla          & 68.77 & 21.16 & 17.45 & 24.21 & 0.38 \\
Summary          & 69.12 & 13.05 &  9.26 & 14.86 & 0.27 \\
Keep-Last-N      & 67.32 &  8.73 &  6.52 &  9.30 & 0.28 \\
LCM              & 66.07 & 17.41 & 32.97 & 18.28 & 0.27 \\
SelectiveContext & 64.02 & 14.60 & 24.95 & 15.25 & 0.29 \\
LLMLingua-2      & 65.13 & 15.80 & 24.68 & 16.82 & 0.30 \\
TokenPilot       & 69.10 & 4.83 & 14.02 & 3.49 & 0.26 \\
\bottomrule[1.2pt]
\end{tabular}
}
\caption{Performance and resource consumption on mixed-category \texttt{PinchBench} streams.}
\label{tab:mixed_overall}
\end{table}

Table~\ref{tab:mixed_overall} reports the full-system comparison. TokenPilot attains a score of 69.10, comparable to Vanilla and Summary, while achieving the lowest cost of \$4.83, with a 44.7\% below Keep-Last-N. The higher absolute cost than in category-grouped streams is expected: adjacent tasks share fewer artifacts and tools, while schema changes shorten reusable prefixes. Nevertheless, ingestion-side reduction remains effective independent of task order, and lifecycle-aware eviction bounds accumulated history, allowing TokenPilot to retain its cost advantage under heterogeneous streams.

\begin{table}[t]
\centering
\setlength{\tabcolsep}{4pt}
\resizebox{\linewidth}{!}{%
\begin{tabular}{l|cc|ccc}
\toprule[1pt]
\textbf{Method} & \textbf{Overall}  & \textbf{Cost (\$)} & \textbf{Miss (M)} & \textbf{Read (M)} & \textbf{Output (M)}\\
\midrule
Vanilla                         & 68.77 & 21.16 & 24.21 & 17.45 & 0.38 \\
\quad + Prefix Stabilization   & 67.83 & 17.09 & 19.02 & 20.33 & 0.29 \\
Summary                         & 69.12 & 13.05 & 14.86 &  9.26 & 0.27 \\
\quad + Prefix Stabilization   & 68.37 & 12.39 & 13.58 & 11.97 & 0.29 \\
Keep-Last-N                     & 67.32 &  8.73 &  9.30 &  6.52 & 0.28 \\
\quad + Prefix Stabilization   & 66.91 &  8.21 &  8.15 &  9.34 & 0.31 \\
\bottomrule[1.2pt]
\end{tabular}
}
\caption{Prefix-stabilized context-management baselines on mixed-category \texttt{PinchBench} streams.}
\label{tab:mixed_prefix_stabilization}
\end{table}

Table~\ref{tab:mixed_prefix_stabilization} shows that prefix stabilization remains beneficial under heterogeneous ordering, reducing cost from \$21.16 to \$17.09 for Vanilla, from \$13.05 to \$12.39 for Summary, and from \$8.73 to \$8.21 for Keep-Last-N. Its smaller gain for fixed-window management highlights the distinction between preserving the invariant core prefix and controlling the mutable task-history suffix: stabilization addresses the former, whereas reduction and lifecycle-aware eviction regulate the latter.

\end{document}